\documentclass[letterpaper, 10pt, conference]{ieeeconf}
\IEEEoverridecommandlockouts 

\usepackage{times}
\usepackage{hyperref}
\usepackage{graphicx}
\usepackage{amsfonts}
\usepackage{amsmath}
\usepackage{amssymb}
\usepackage{array}
\usepackage{flafter}
\usepackage{cite}
\usepackage{subfigure}
\usepackage{verbatim}
\usepackage{multirow}
\usepackage[usenames,dvipsnames]{color}
\usepackage[ruled,vlined,linesnumbered]{algorithm2e}

\newcommand{\graph}{\mathcal{G}}
\newcommand{\locallayer}{\mathcal{L}_{\rm local}}
\newcommand{\globallayer}{\mathcal{L}_{\rm global}}

\newcommand{\globalgraph}{\graph}

\newcommand{\point}{\mathbf{p}}

\newcommand{\pointset}{\mathcal{P}}

\title{\huge FAR Planner: Fast,  Attemptable Route Planner \\ using Dynamic Visibility Update
\thanks{All authors are with CMU Robotics Institute. Emails: {\tt \{fanyang2, ccao1, hongbiaz, jeanoh, zhangji\}@cmu.edu}}}
\author{Fan Yang, Chao Cao, Hongbiao Zhu, Jean Oh,  and Ji Zhang}

\begin{document}

\maketitle

\begin{abstract}
The problem of path planning in unknown environments remains a challenging problem - as the environment is gradually observed during the navigation, the underlying planner has to update the environment representation and replan, promptly and constantly, to account for the new observations. In this paper, we present a visibility graph-based planning framework capable of dealing with navigation tasks in both known and unknown environments. The planner employs a polygonal representation of the environment and constructs the representation by extracting edge points around obstacles to form enclosed polygons. With that, the method dynamically updates a global visibility graph using a two-layered data structure, expanding the visibility edges along with the navigation and removing edges that become occluded by newly observed obstacles. When navigating in unknown environments, the method is attemptable in discovering a way to the goal by picking up the environment layout on the fly, updating the visibility graph, and fast re-planning corresponding to the newly observed environment. We evaluate the method in simulated and real-world settings. The method shows the capability to attempt and navigate through unknown environments, reducing the travel time by up to 12-47\% from search-based methods: A*, D* Lite, and more than 24-35\% than sampling-based methods: RRT*, BIT*, and SPARS.
\end{abstract}

\begin{figure}[t]
	\vspace{0.08in}
    \centering
    \includegraphics[width=0.95\linewidth]{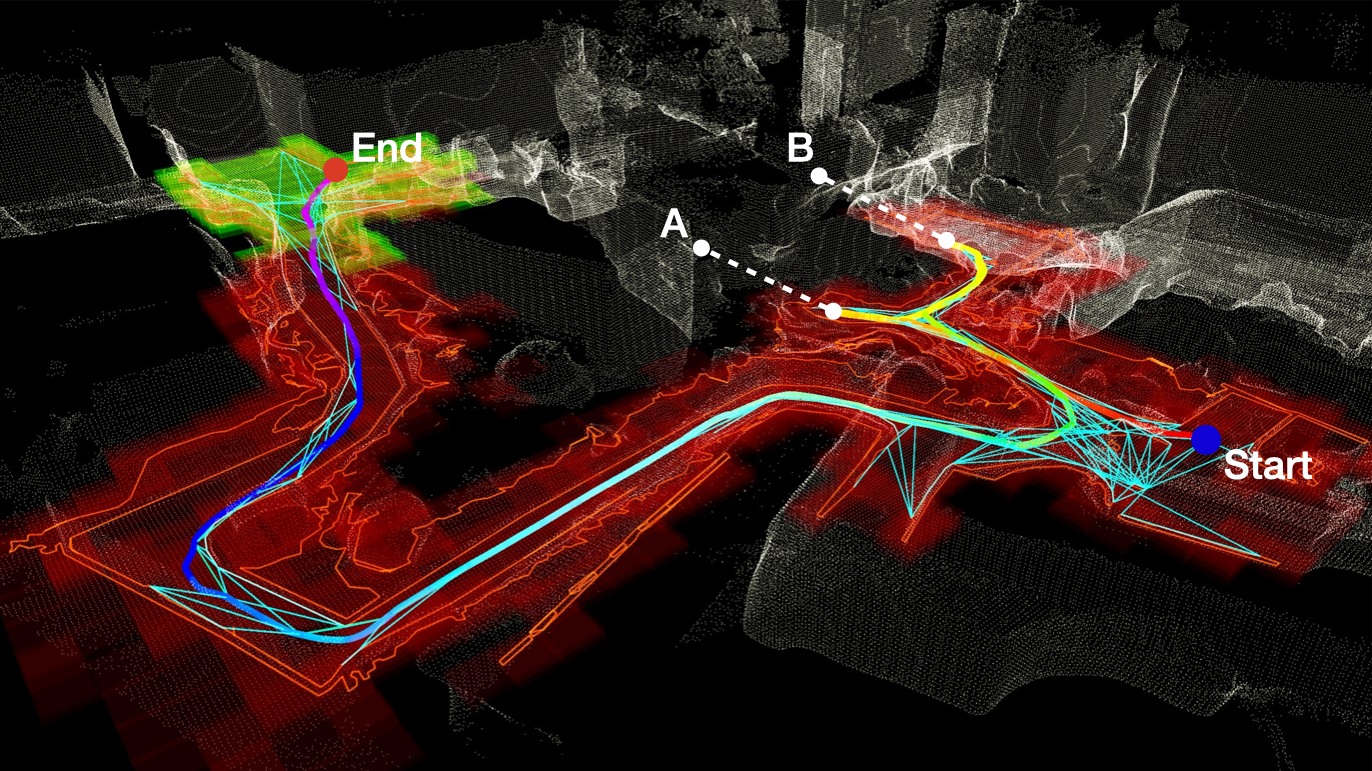}
	\vspace{-0.05in}
    \caption{Navigation through an unknown environment based on a simulated campus model. The colorful curve is the vehicle trajectory staring at the blue dot and ending at the red dot. The planner attempts to guide the vehicle to the goal by registering obstacles in the environment (red polygons) and building a visibility graph (cyan lines) along with the navigation. The red area shows the global environment observed, and the yellow area represents the local environment used to update the local graph. A and B are dead-ends. The vehicle first attempts the path that leads to the dead-ends, then re-plans to guide the vehicle out and finally reaches the goal.}
	\label{fig:opening_fig}
	\vspace{-0.2in}
\end{figure}


\section{Introduction}
Visibility graph-based planning has been studied by the research society but has not gained substantial popularity. The main difficulty in visibility graph-based planning has to do with its requirement on polygonal world \cite{10.1145/359156.359164}. Very often, a considerable amount of computation has to be involved in constructing the visibility graph \cite{Kitzinger2003TheVG}, especially if the environment is complex and three-dimensional (3D). In this paper, we reconsider visibility graph in solving the path planning problem and demonstrate its strength in fast re-planning and the ability to handle unknown and partially known environments. Our method benefits from the fact that visibility edges connect between obstacles. When navigating in an unknown environment, the unseen areas contain few obstacles, involving a small number of visibility edges and hence a low computational cost to adjust the visibility graph -- such adjustments often happen repetitively as more areas are observed along with the navigation.



Our method employs a two-layered framework for real-time visibility graph construction. On a local layer, the method constructs the visibility graph at every data frame. This uses data acquired from range sensors, from which, the method extracts edge points around the obstacles and converts the edge points into a set of enclosed polygons. Then, visibility edges are connected between the polygons to form the visibility graph, which is merged into a global layer and maintained at the global scale. The incremental graph construction results in a low computational cost needing only a small amount of processing ($\sim$20\% of a single i7 CPU thread). For aerial vehicle planning in a 3D environment, the method is further extended to incorporate a multi-layer polygonal representation.

With the visibility graph dynamically built and maintained, the method searches the graph for the shortest path at low latency (within $\sim$10ms), resulting in a fast response after receiving a goal. When navigating through unknown environments, the method constantly attempts multiple routes in a sequential manner to guide the vehicle in finally reaching the goal. In the case that dynamic obstacles are present in the environment, the method eliminates visibility edges blocked by the dynamic obstacles and later on reconnects the edges after regaining the visibility.

We evaluate our method using both ground and aerial vehicles, in simulated and real-world experiments. Our ground vehicle simulation environments include a moderately convoluted indoor environment, a mid-scale outdoor campus environment, and a large-scale, highly convoluted tunnel-network environment. We benchmark the performance of the state-of-the-art planners in handling navigation tasks through unknown or partially known environments and compare them with our method. We conclude that in large-scale, highly convoluted environments, our method outperforms the state-of-the-art planners in the planning time and the time to travel to goals. 

The main contributions of this paper are summarized as follows:
\begin{itemize}
  \item A two-layered algorithm framework for polygon extraction from obstacles and incremental visibility graph construction at a low computational cost.
  \item The framework is capable of dynamically adjusting the visibility graph for attemptable navigation in unknown environments and for handling dynamic obstacles.
  \item Benchmarking state-of-the-art planner performance in unknown and partially known environments.
\end{itemize}

The FAR Planner open-source software\footnote{FAR Planner: \url{github.com/MichaelFYang/far_planner}} has been integrated to our open-source Autonomous Exploration Development Environment \cite{Chao2022DevEnv}\footnote{Development Environment: \url{www.cmu-exploration.com}} to promote research in navigation autonomy. 
The two repositories form a full stack of planning algorithms for ground vehicle navigation.

\section{Related Work}

The path planning problem has been tackled from multiple angles. However, navigation and planning through unknown environments still remain challenging especially in real-world settings. The approach described here is based on key results of random sampling-based, search-based, and learning-based planners, solving the planning problem in both known and unknown environments.


\textit{Random sampling-based planners}: The classic methods in the Rapidly-expanding Random Tree (RRT) \cite{lavalle2001rapidly} family include the original RRT and its variances such as RRT* \cite{karaman2011sampling}, RRT-Connect \cite{844730}, Informed RRT* \cite{6942976}, and BIT* \cite{bitrrt2038457}. Together with probabilistic roadmap-based (PRM) methods \cite{508439} such as Lazy PRM \cite{844107}, and the latest SPARS \cite{6631156} which produces sparse subgraphs for fast query resolution, these methods excel at exploring free space in the environments. However, without a prior map, those methods suffer from a long planning time in order to draw samples from both free and unknown space. Approaches \cite{4269896, 1242258} are developed to draw samples biased to the goal or reuse previous samples \cite{1641879} to reduce the redundant sampling in unknown space. Those methods are either greedy-guided which can be easily trapped by local minima or require high computational cost to maintain the tree/graph from previous iterations to account for the newly observed environment.


\textit{Search-based planners}: These methods include Dijkstra's algorithm \cite{dijkstra1959note}, A* \cite{Hart1968}, D* \cite{stentz1997optimal}, and D* Lite \cite{koenig2005fast}. Dijkstra's algorithm and A* search on a discretized grid. The methods often need a long planning time because they reinitialize the propagation at each planning cycle. In addition, for unknown environments, those methods often require a predefined region for discretization and are not scalable. The incremental versions of A*: D* and D* Lite, are proposed to efficiently handle environment changes for navigation through unknown terrains. The methods reduce planning time by reusing the results from the previous planning cycle and adjusting only the local inconsistent states. However, recent work \cite{7926533, 5679403} shows that when the vehicle reaches a dead-end and requires a distinctive path from the previous plan to leave the dead-end, the planning time can be significant and even surpass the non-incremental versions.

\textit{Learning-based planners}: These methods \cite{Bayesian43242, learnNav24i3204, 9066637} need to be trained by a supervising method or using ground-truth data. The training process essentially encodes map information in the internal representation, e.g., a deep network. At test time, the methods can handle environments sharing similar settings with the training environments. In essence, learning-based planners are data-driven and can be limited to the environments that are present in the training data.

This paper focuses on a metric-based method with dynamic visibility graph updates. Although using visibility graph for robot navigation has been studied in the literature \cite{10.1145/359156.359164, 1087133, 384257, 1642054, 5967147, Visibility118.123}, it is not well applied in real-world applications due to its high computation complexity and the requirement for well-defined polygonal geometry. Our work adapts the visibility graph-based method and benefits from its line-of-sight, relatively sparse visibility edges for fast planning in unknown and partially known environments.

\begin{figure}[t!]
    \centering
    \subfigure[]{\includegraphics[height=0.45\linewidth]{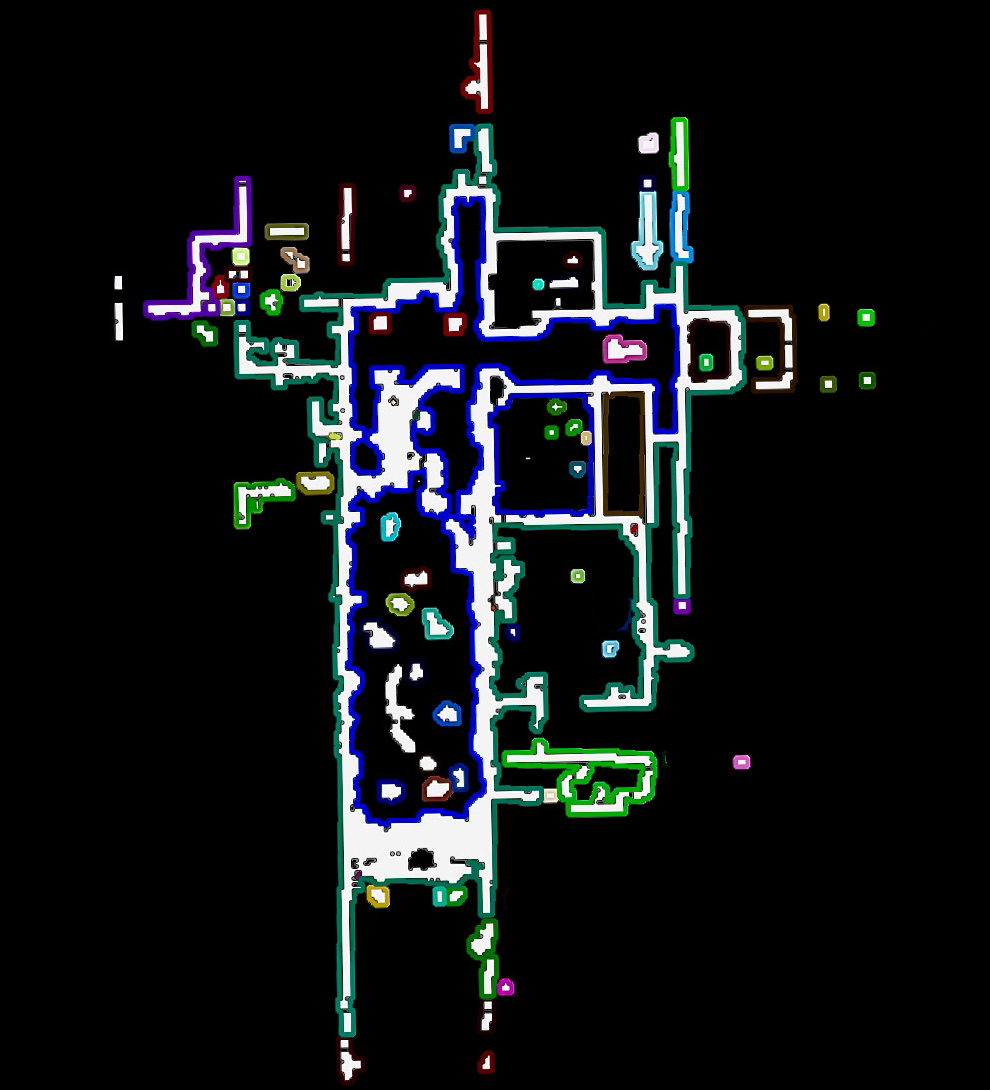}} \hspace{0.05in}
    \subfigure[]{\includegraphics[height=0.45\linewidth]{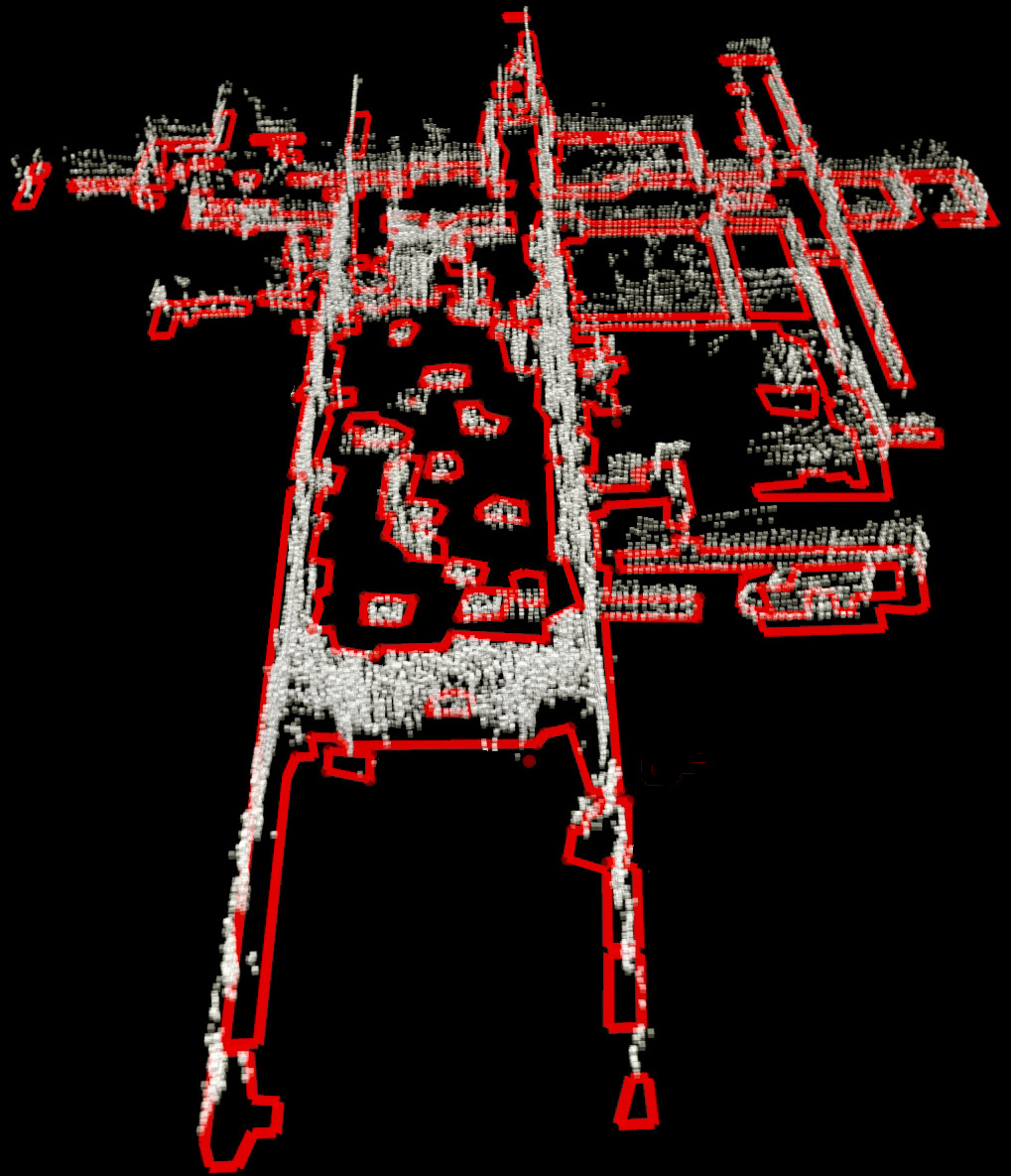}} \\\vspace{-0.1in}
    \caption{(a) Polygons $\{\pointset_{\rm contour}^{k}\}$ extracted from blurred image $\mathcal{I_{\rm blur}}$ with dense vertices. The black pixels are traversable and the white pixels are obstacles. Different colors represent different polygons. (b) Final, extracted polygons $\{\pointset_{\rm local}^{k}\}$ (red) collated with sensor data points $\mathcal{S}$ (white).}
	\label{fig:polygon}
	\vspace{-0.2in}
\end{figure}

\section{Methodology}

Define $\mathcal{Q} \subset \mathbb{R}^3$ as the work space for the robot to navigate. Let $\mathcal{S} \subset \mathcal{Q}$ be the set of sensor data points from obstacles. Our method develops a visibility graph (v-graph), denoted as $\graph$, from $\mathcal{S}$. Given the robot position $\point_{\rm robot} \in \mathcal{Q}$ and goal $\point_{\rm goal} \in \mathcal{Q}$, a path can be searched between $\point_{\rm robot}$ and $\point_{\rm goal}$.

\subsection{Obstacle Polygon Extraction and Registration}

We describe the process of converting sensor data points $\mathcal{S}$ into a set of polygons, denoted as $\{\pointset_{\rm local}^{k} \subset \mathcal{Q}~|~k\in\mathbb{Z}^+\}$. For ground vehicles, very often, a terrain traversability analysis module runs in the system to analyze the terrain characteristics. The module takes in range measurements such as from Lidar or depth camera and outputs $\mathcal{S}$ representing the obstacles. The polygon extraction process uses image processing techniques. Let $\mathcal{I}$ be a binary image where a black pixel corresponds to a point in the traversable space and a white pixel corresponds to a point on an obstacle, $\mathcal{I}$ is centered at the robot position $\point_{\rm robot}$. We first project $\mathcal{S}$ onto $\mathcal{I}$ and at the same time inflate the points in $\mathcal{S}$ using the vehicle size. Then, we blur the image with an average filter to create a grayscale image $\mathcal{I}_{\rm blur}$. After that, we extract edge points in $\mathcal{I}_{\rm blur}$ and analyze the topological distribution of the edge points using the method in \cite{Suzuki1985TopologicalSA}. This gives us a set of enclosed polygons with dense vertices along the contour, as shown in Fig.~\ref{fig:polygon}(a). Denote the polygons as $\{\pointset_{\rm contour}^{k} \subset \mathcal{Q}~|~k\in\mathbb{Z}^+\}$. For each $\pointset_{\rm contour}^{k}$, we use the method in \cite{Douglas1973ALGORITHMSFT} to downsize the vertices and further check the inner angle between the two connected edges on the contour for each vertex to infer the local curvature of the obstacle. The vertices with the inner angle less than a threshold are eliminated. The final, extracted polygons $\{\pointset_{\rm local}^{k}\}$ are shown in Fig.~\ref{fig:polygon}(b). To help readers follow the process, we write down the polygon extraction algorithm in Algorithm 1.

\begin{figure}[t]
    \centering
    \includegraphics[width=0.7\linewidth]{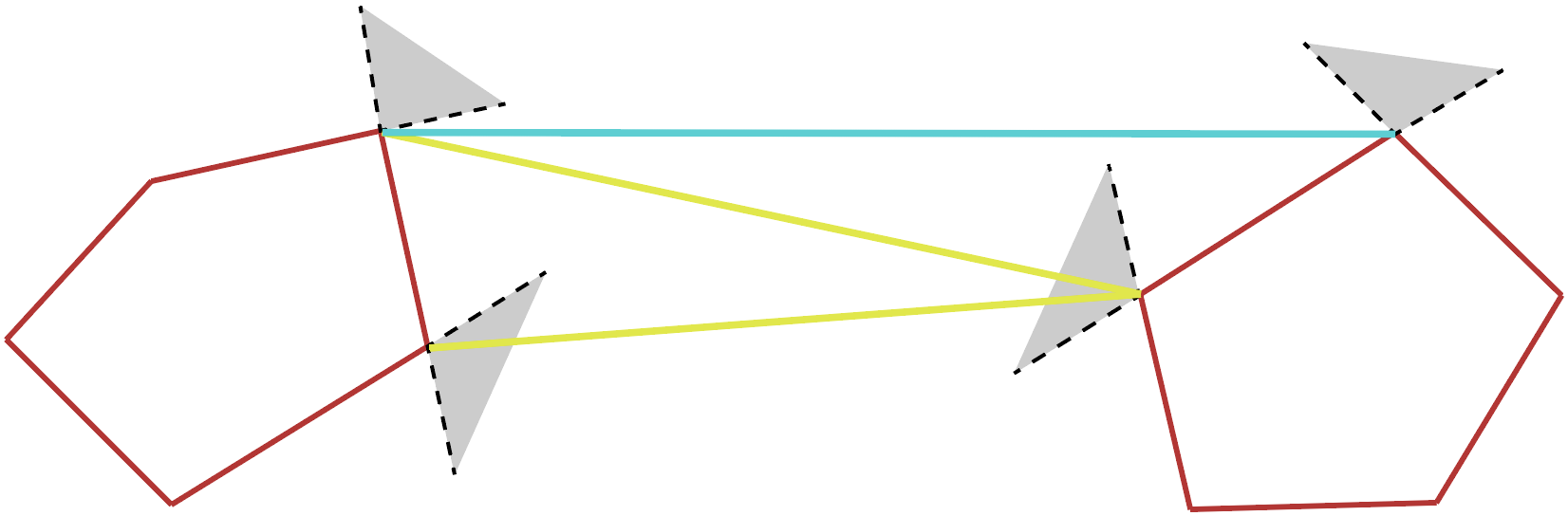}
	\vspace{-0.05in}
    \caption{Illustration of reduced v-graph. The two red polygons represent $\{\pointset_{\rm local}^{k}\}$. Visibility edges (yellow) that head into one or both connected polygons from the shaded angles are useless and eliminated, and edges (cyan) that ``pass around" the polygons are kept.}
	\label{fig:reduced}
	\vspace{-0.15in}
\end{figure}


\begin{algorithm}[b!]
\caption{Polygon Extraction and Registration}
\label{algo:polygon_extraction}
\SetAlgoLined
\LinesNumbered
\SetKwInOut{Input}{Input}
\SetKwInOut{Output}{Output}
\Input {Sensor Data Points: $\mathcal{S}$}
\Output {Polygons: $\{\pointset_{\rm local}^{k}\}$}
Create binary image $\mathcal{I}$ from points in $\mathcal{S}$\;
Apply average filter to generate blurred image $\mathcal{I_{\rm blur}}$\;
Extract polygons $\{\pointset_{\rm contour}^{k}\}$ based on \cite{Suzuki1985TopologicalSA}\;
\For{{\rm each} $\pointset_{\rm contour}^{k}$} {
    Downsample vertices in $\pointset_{\rm contour}^{k}$ based on \cite{Douglas1973ALGORITHMSFT}\;
    Check the inner angle of each vertex and eliminate the vertices with inner angle $<\zeta$\; 
    Use vertices kept to form final polygon $\pointset_{\rm local}^{k}$\;
}
\end{algorithm}

\subsection{Two-layer V-graph Dynamic Update}

The v-graph $\graph$ employed in this paper contains two layers -- a local layer, denoted as $\mathcal{L}_{\rm local}$, surrounding the robot and a global layer, denoted as $\mathcal{L}_{\rm global}$, covering the observed environment. At each data frame, $\mathcal{L}_{\rm local}$ is generated from sensor data points $\mathcal{S}$ and then merged with $\mathcal{L}_{\rm global}$. We know that the computational complexity of constructing a v-graph is $O(n^2\log n)$ \cite{Lee78}, where $n$ is the number of vertices on the v-graph. Given that our v-graph is constructed locally on $\mathcal{L}_{\rm local}$, the computational cost is considerably limited. In other words, our method incrementally updates the v-graph to distribute the computation evenly to every data frame by updating only the area in the vicinity of the vehicle.


\textit{Constructing Local Layer}: Recall that the sensor data points $\mathcal{S}$ are converted into polygons $\{\pointset_{\rm local}^{k}\}$. With $\{\pointset_{\rm local}^{k}\}$, we construct a partially reduced v-graph on the local layer $\mathcal{L}_{\rm local}$.
Let $\mathcal{E}_{\rm local}$ be the set of visibility edges on $\mathcal{L}_{\rm local}$. Specifically, for the edges in $\mathcal{E}_{\rm local}$ that are longer than a threshold, we neglect the unnecessary edges that head into the one or both connected polygons and keep the edges that ``pass around" (see Fig.~\ref{fig:reduced}). The edges in $\mathcal{E}_{\rm local}$ that are shorter than the threshold are all kept without ``reduction". This is due to the effect of position noises of the vertices in $\mathcal{E}_{\rm local}$ is increased with shorter edges, causing larger direction changes of the edges and making it hard to identify whether the edges are heading into the polygon or not. Specifically, the edges that form the polygons in $\{\pointset_{\rm local}^{k}\}$ are also kept in $\mathcal{E}_{\rm local}$ as they ``pass around" the polygons. In practice, we observe that the traversable space in an environment is often constrained where majority of the visibility edges are blocked by polygons close by, resulting in a relatively small number of final, connected edges in $\mathcal{E}_{\rm local}$.

\begin{algorithm}[t!]
\caption{Dynamic V-graph Update}
\label{algo:graph_update}
\SetAlgoLined
\LinesNumbered
\SetKwInOut{Input}{Input}
\SetKwInOut{Output}{Output}
\Input {Sensor Data: $\mathcal{S}$, V-graph: $\globalgraph$}
\Output {Updated v-graph $\globalgraph$}
$\{\pointset_{\rm vertex}^{k}\} \leftarrow$ PolygonExtraction($\mathcal{S}$)\;
Construct partially reduced v-graph on $\mathcal{L}_{\rm local}$\;
Associate vertices between $\{\pointset_{\rm local}^{k}\}$ and $\{\pointset_{\rm global}^{l}\}$\;
\For{{\rm each vertex} $\in \{\pointset_{\rm local}^{k}\} \bigcup \{\pointset_{\rm global}^{l}\}$} {
  \If{{\rm an association exists}} {
    Update the vertex in $\{\pointset_{\rm global}^{l}\}$ to the mean of the inliers assigned by robust fitting\;
  }
  \ElseIf{{\rm the vertex} $\in \{\pointset_{\rm global}^{l}\}$}{
    Remove the vertex from $\{\pointset_{\rm global}^{l}\}$ based on voting result\;
  }\Else{
    Add the vertex to $\{\pointset_{\rm global}^{l}\}$ as a new vertex\;
  }
}
Merge edges from $\mathcal{E}_{\rm local}$ into in $\mathcal{E}_{\rm global}$ and eliminate those blocked or connected with removed vertices\;
\end{algorithm}

\textit{Updating Global Layer}: After constructing the local v-graph on $\locallayer$, we merge $\locallayer$ with $\globallayer$ to update $\globallayer$ in the area overlapping with $\locallayer$. Define $\{\pointset_{\rm global}^{l} \subset \mathcal{Q}~|~l\in\mathbb{Z}^+\}$ as the set of polygons and $\mathcal{E}_{\rm global}$ as the set of visibility edges on $\globallayer$. The process starts with associating the vertices between $\{\pointset_{\rm local}^{k}\}$ and $\{\pointset_{\rm global}^{l}\}$. A vertex in $\{\pointset_{\rm local}^{k}\}$ is associated to the vertex in $\{\pointset_{\rm global}^{l}\}$ only if they are the closest vertex to each other and the euclidean distance in between is less than a threshold. Then, for the vertices in $\{\pointset_{\rm global}^{l}\}$ that are associated, the positions are updated. Here, we use robust fitting \cite{andersen2008modern} to further eliminate outlier associations. Given a vertex in $\{\pointset_{\rm global}^{l}\}$, the corresponding vertices over a number of data frames are filtered through an iteration process. The iterations start with all the vertices as inliers. At each iteration, we re-calculate the mean and covariance of the inliers and use those to re-assign the inliers and outliers among the vertices. The iterations terminate if the inliers stay the same over two consecutive iterations or a maximum iteration number is meet. Then, the vertex in $\{\pointset_{\rm global}^{l}\}$ is updated to the mean of the inliers. For the vertices in $\{\pointset_{\rm global}^{l}\}$ that are not associated, they are removed based on a voting result where an association is not found for certain times over a number of data frames. For the vertices in $\{\pointset_{\rm local}^{k}\}$ that are not associated, they are added to $\{\pointset_{\rm global}^{l}\}$ as new vertices.
Finally, the edges in $\mathcal{E}_{\rm local}$ are merged into $\mathcal{E}_{\rm global}$. If the edges exist in $\mathcal{E}_{\rm global}$, they are updated, otherwise they are added as new edges. The overall process of the v-graph update is presented in Algorithm 2.

\begin{figure}[b!]
	\vspace{-0.1in}
    \centering
    \includegraphics[width=0.95\linewidth]{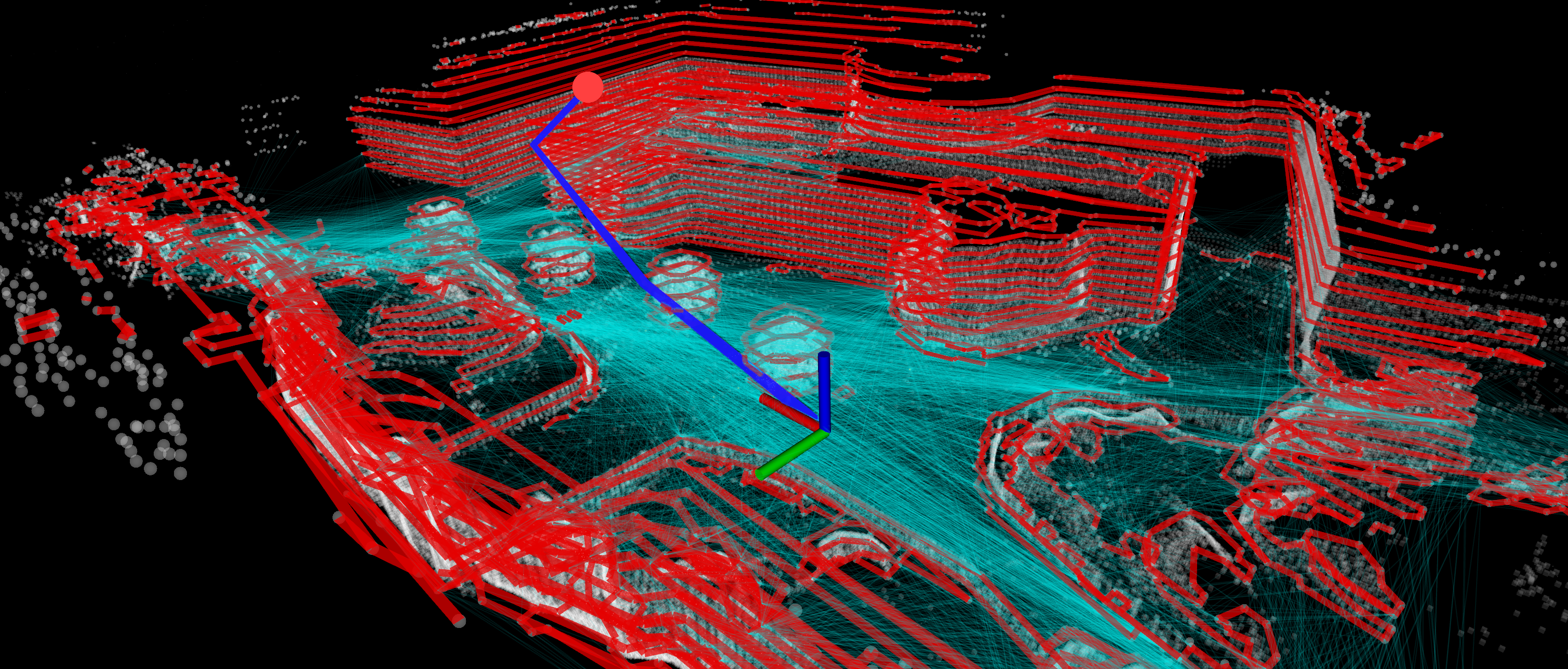}
    \vspace{-0.05in}
    \caption{Example 3D multi-layer v-graph. Multi-Layer polygons (red) are extracted from a 3D environment. The visibility edges (cyan lines) cross multiple polygon layers. Further, a path (blue) is searched on the 3D multi-layer v-graph between the vehicle (coordinate frame) and goal (red dot).}
	\label{fig:3D_V-grapg}
\end{figure}

\subsection{Planning on V-Graph}

Given the robot position $\point_{\rm robot}$ and goal $\point_{\rm goal}$, we would like to search the v-graph $\globalgraph$ for the shortest path between $\point_{\rm robot}$ and $\point_{\rm goal}$. The planner first add $\point_{\rm robot}$ and $\point_{\rm goal}$ as two vertices on the global layer $\globallayer$ and connect them to the vertices in $\{\pointset_{\rm global}^{l}\}$ with non-blocking visibility edges. Then, a breath-first search is run on $\globallayer$ to propagate through $\mathcal{E}_{\rm global}$ and find the shortest path between $\point_{\rm robot}$ and $\point_{\rm goal}$, if a path is available.

As the robot navigates through the environment, the vertices in $\{\pointset_{\rm global}^{l}\}$ that have established non-blocking visibility edges with $\point_{\rm robot}$ form the free space, and the rest vertices in $\{\pointset_{\rm global}^{l}\}$ form the unknown space. After the navigation completes, the v-graph is saved with a label associated with each vertex to indicate the type of space. For future runs, the v-graph can be loaded into the planner as a prior map. When searching for a path through the vertices, we provide the option of searching in the combined space (free and unknown) for attemptable planning or in free space only for non-attemptable planning.




\subsection{Extension to 3D Multi-Layer V-Graph}

An extended 3D version of our method for aerial vehicle planning models the environment as multiple horizontal slices and extracts multi-layer polygons. The visibility edges are connected across multiple polygon layers. Note that the partially reduced mechanism only applies to the visibility edges on a single polygon layer. For the visibility edges that connect different polygon layers, we keep all the non-blocking edges. Also, for the visibility edge that crosses three or more polygon layers, the collision check takes into account the blockage of the polygons on a mid-layer that the edge passes through. An example 3D v-graph and a path searched on the v-graph are shown in Fig.~\ref{fig:3D_V-grapg}.


\begin{figure}[b!]
    \vspace{-0.15in}
	\centering
	\subfigure[]{\includegraphics[height=0.35\linewidth]{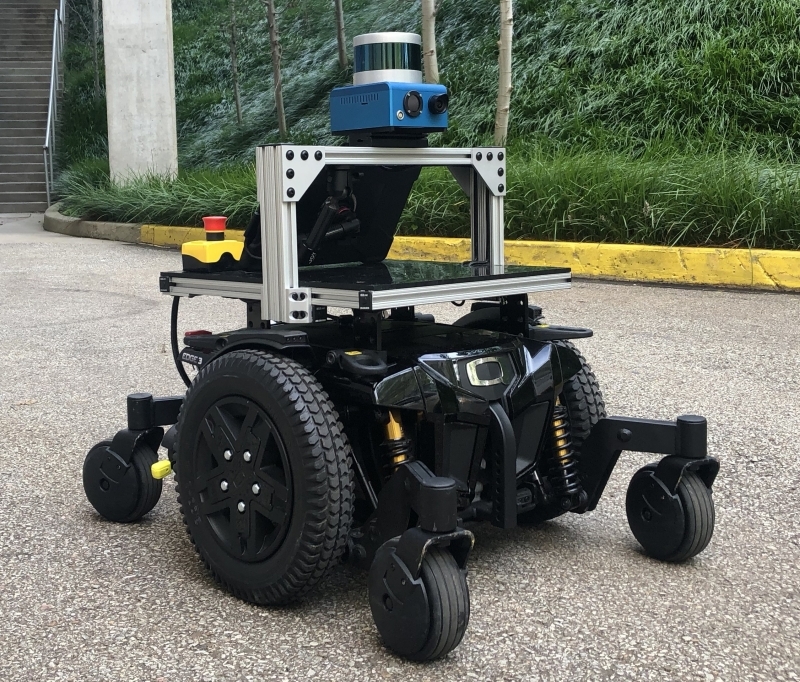}} \hspace{0.1in}
	\subfigure[]{\includegraphics[height=0.35\linewidth]{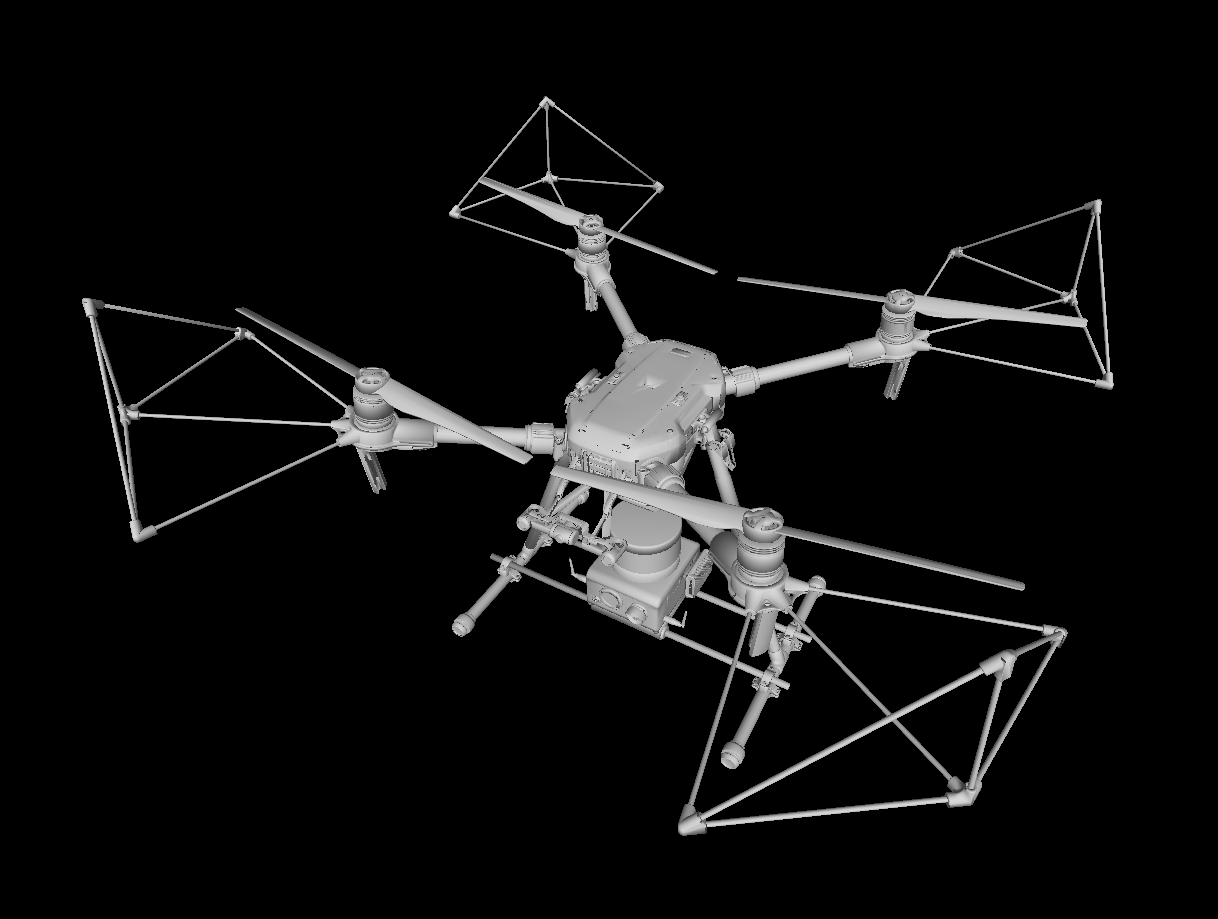}} \\\vspace{-0.1in}
	\caption{(a) Ground experiment and (b) Simulated aerial platforms.}
	\label{fig:vehicles}
\end{figure}

\section{Experiments}

Our ground vehicle platform and simulated aerial vehicle platform are shown in Fig. \ref{fig:vehicles}. Both vehicles are equipped with a Velodyne Puck Lidar used as the range sensor for navigation planning. The ground vehicle has a camera at $640 \times 360$ resolution and a MEMS-based IMU, coupled with the Lidar for state estimation \cite{zhang2018slam}. The onboard autonomy system incorporates navigation modules from our development environment, e.g., terrain analysis, and way-point following based on kinodynamic feasible trajectories generated by the local planner \cite{zhang2020avoidance}, as fundamental navigation modules and runs FAR planner at the top of the system.

In the experiments, we compare our method to two search-based methods: \textbf{A*}\cite{Hart1968}, \textbf{D* Lite}\cite{koenig2005fast}, and three sampling-based methods: \textbf{RRT*}\cite{karaman2011sampling}, \textbf{BIT}*\cite{bitrrt2038457}, and \textbf{SPARS}\cite{6631156}. Here, BIT* is considered the state-of-the-art in the RRT-based family and RRT* is the classic method of the family. SPARS is considered the state-of-the-art in the PRM-based family. All methods run on a 4.1GHz i7 computer. We configure FAR planner to update the v-graph at 2.5Hz and perform a path search for re-planning at each v-graph update. The planner uses images at 0.2m/pixel resolution to extract edge points for constructing polygons. The local layer on the v-graph is a 40m$\times$40m area with the vehicle in the center. 

\begin{figure}[t!]
	\centering
	\subfigure[]{\includegraphics[width=0.88\linewidth]{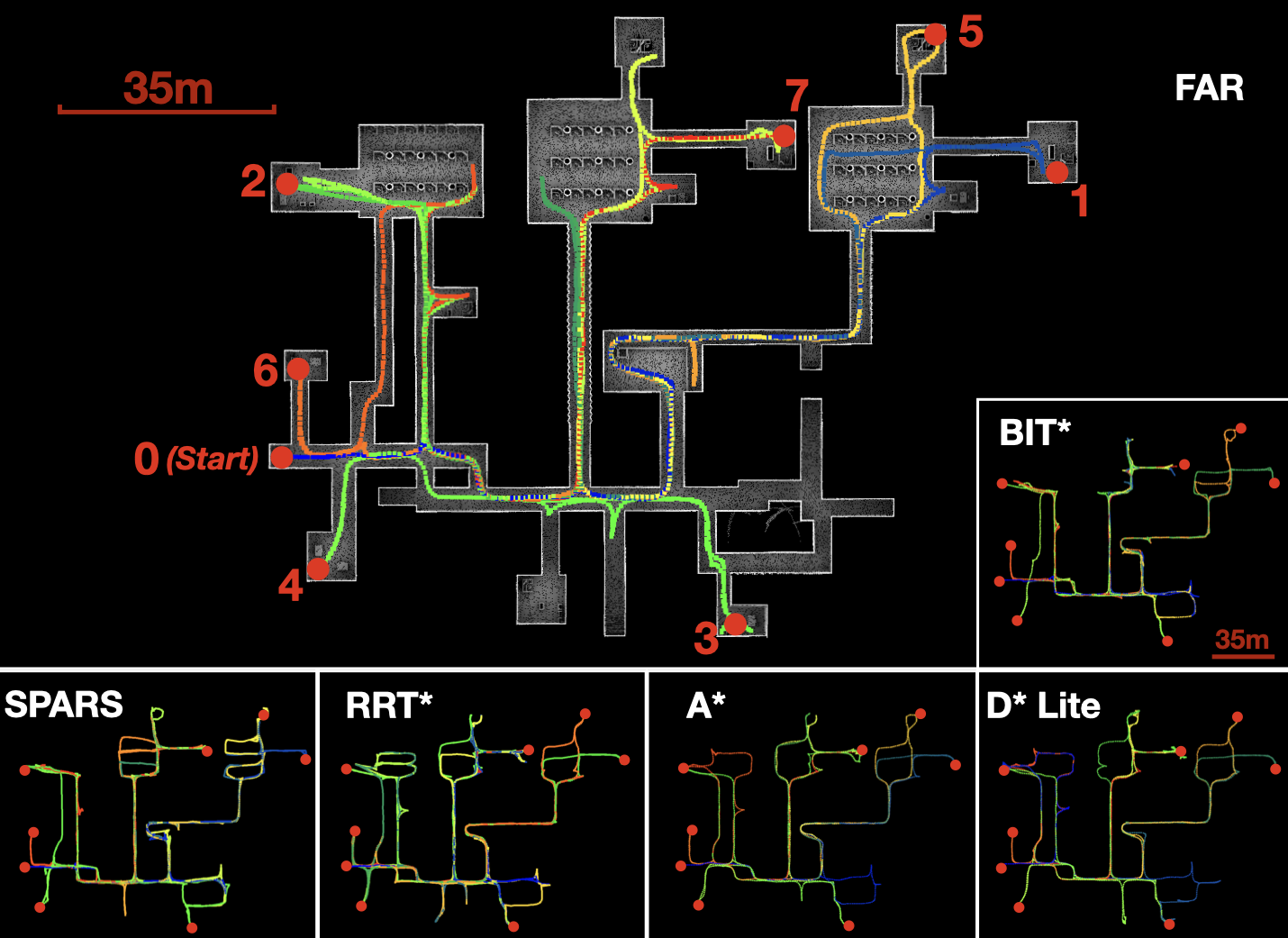}}\\\vspace{-0.1in}
	\subfigure[]{\includegraphics[width=0.79\linewidth]{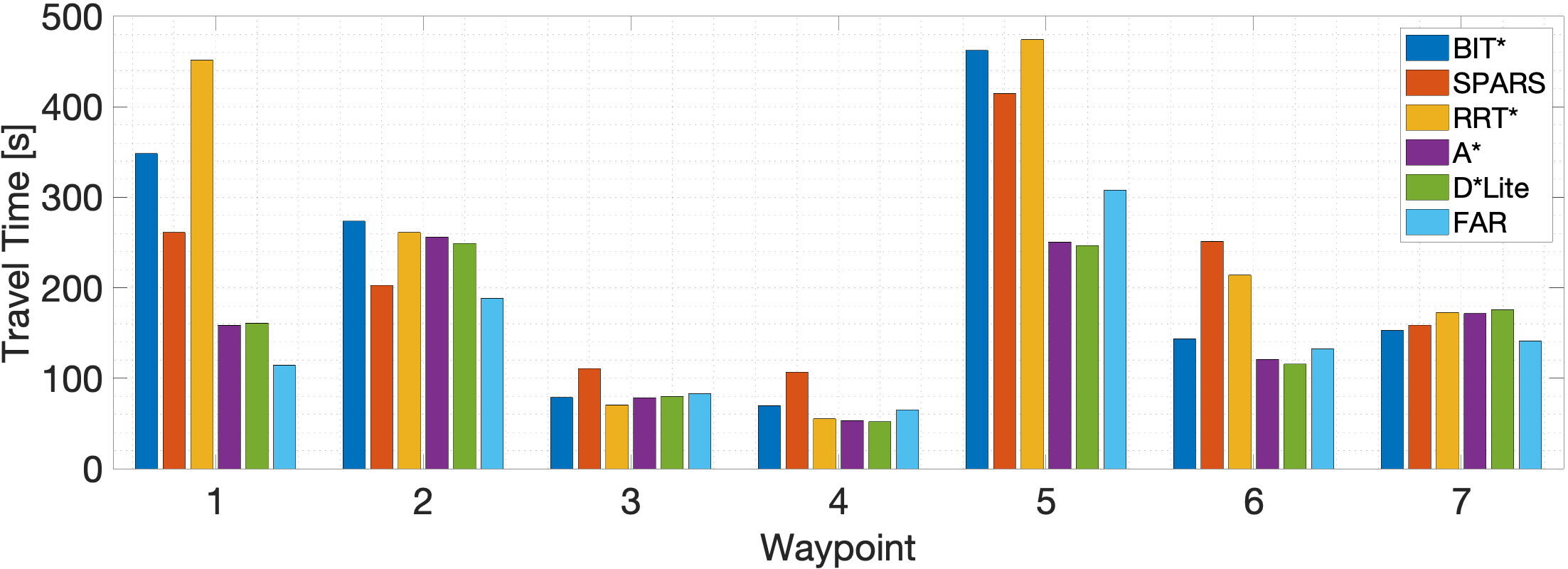}}\\\vspace{-0.1in}
	\subfigure[]{\includegraphics[width=0.79\linewidth]{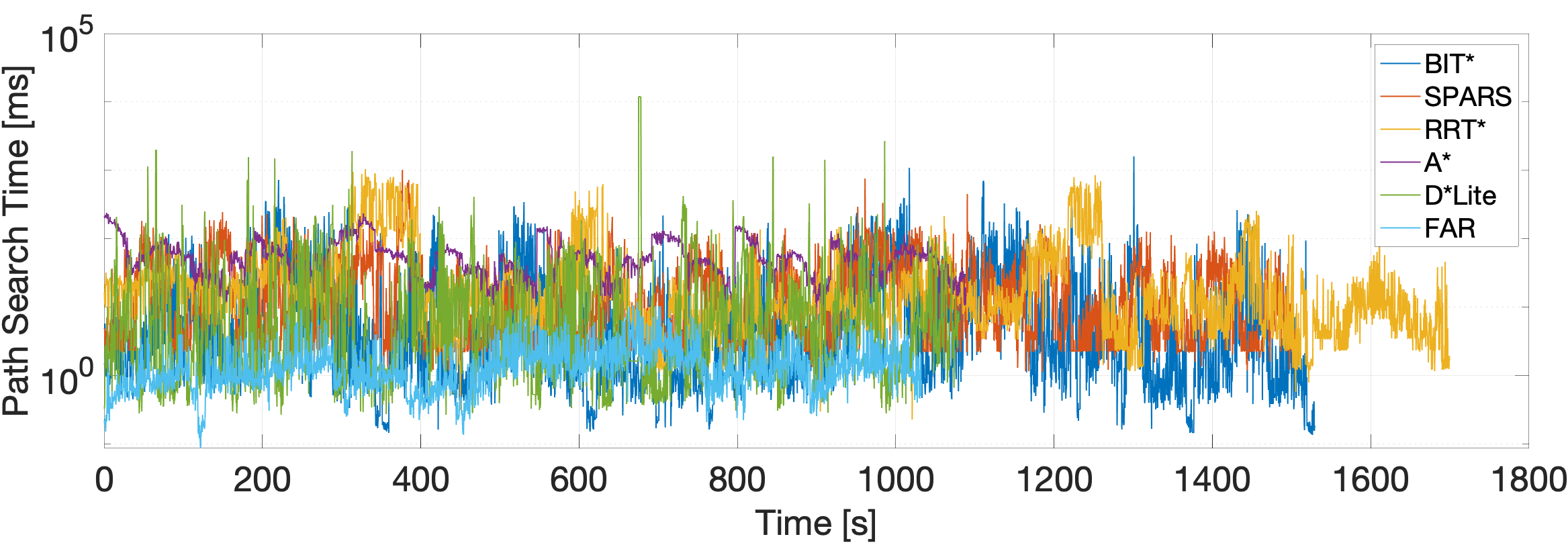}}\\\vspace{-0.1in}
	\subfigure[]{\includegraphics[width=0.79\linewidth]{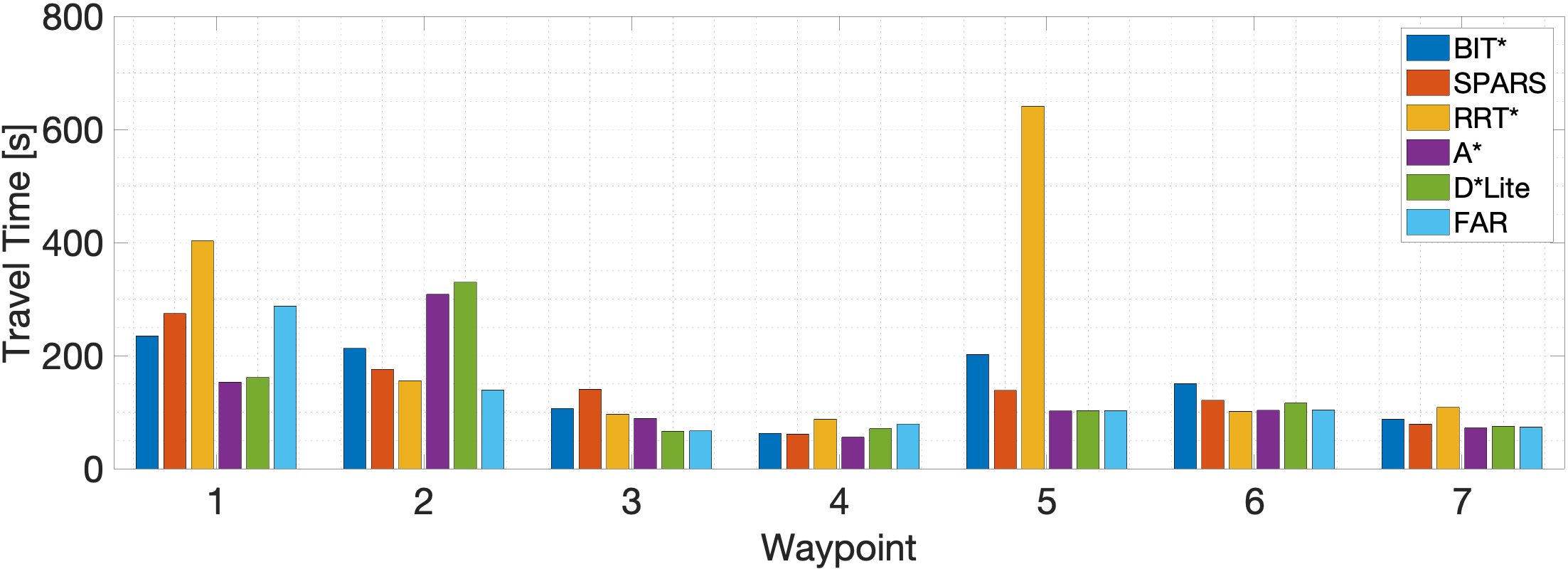}}\\\vspace{-0.1in}
	\subfigure[]{\includegraphics[width=0.79\linewidth]{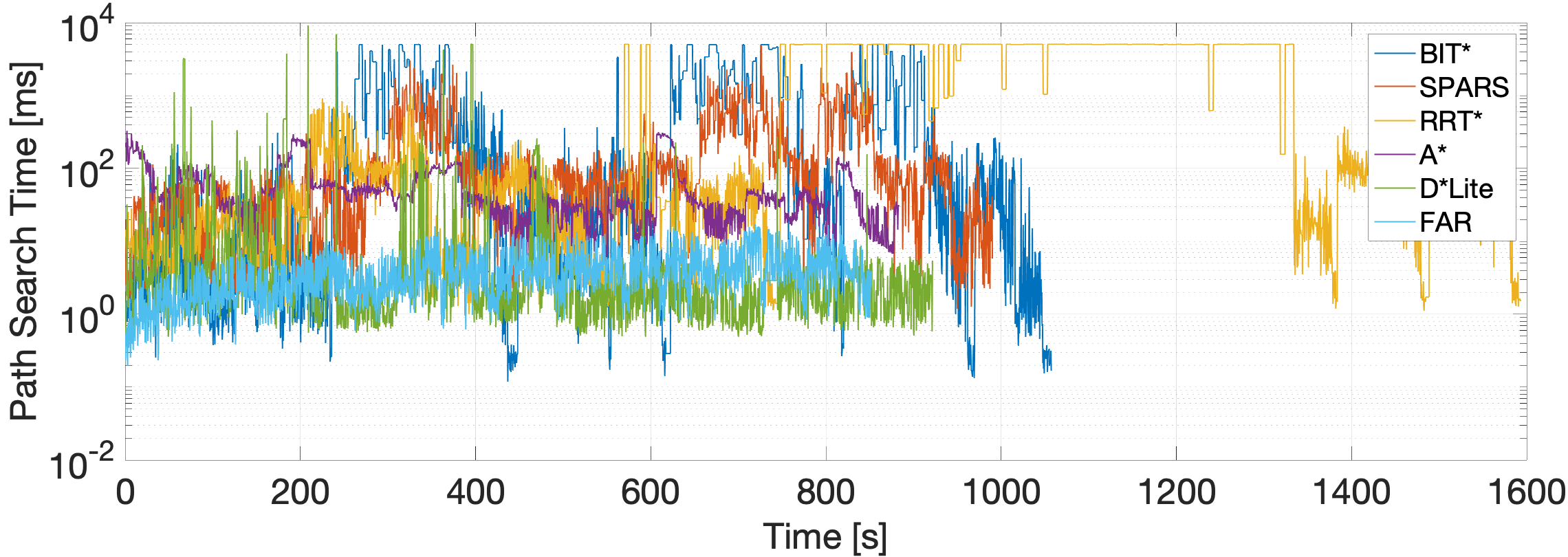}}\\\vspace{-0.1in}
	\caption{Ground vehicle simulation in mid-scale, moderately convoluted indoor environment. The vehicle is set to navigate in increasing order from the start (point 0) to point 7 (see labels in (a)). The experiment is conducted in two runs. In the first run, the planner is reset after arriving at each point. This mimics navigation in an unknown environment. In the second run, the planner accumulates the environment observations through the run. This mimics navigation in a partially known environment. The trajectories in (a) are from the run with planner reset. (b) and (c) present the time of arrival at each point and the search time with planner reset. (d) and (e) show the same metrics with accumulated environment observations.}
	\label{fig:indoor}
    \vspace{-0.2in}
\end{figure}

\begin{figure}[t!]
	\centering
	\subfigure[]{\includegraphics[width=0.865\linewidth]{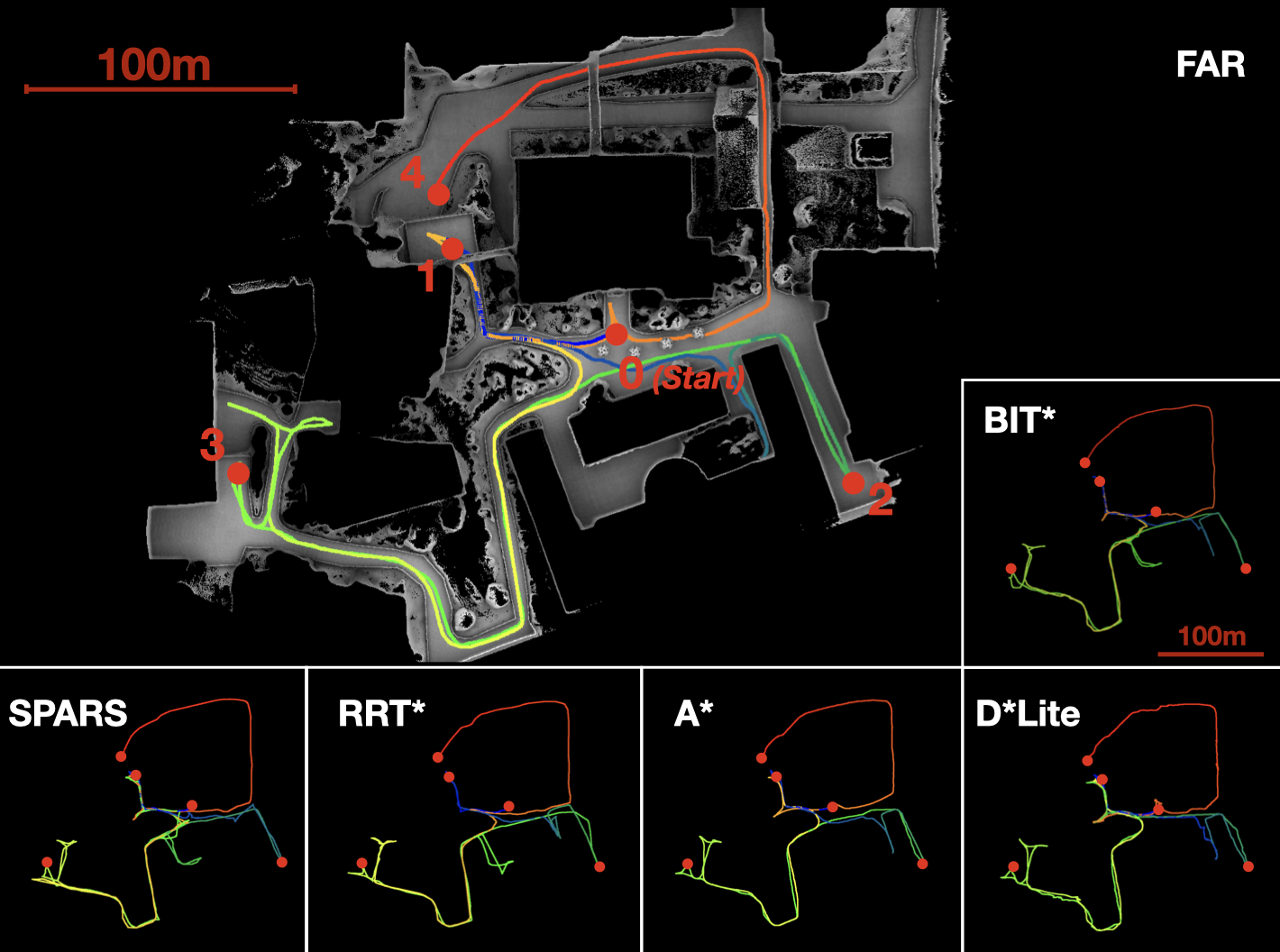}}\\\vspace{-0.1in}
	\subfigure[]{\includegraphics[width=0.79\linewidth]{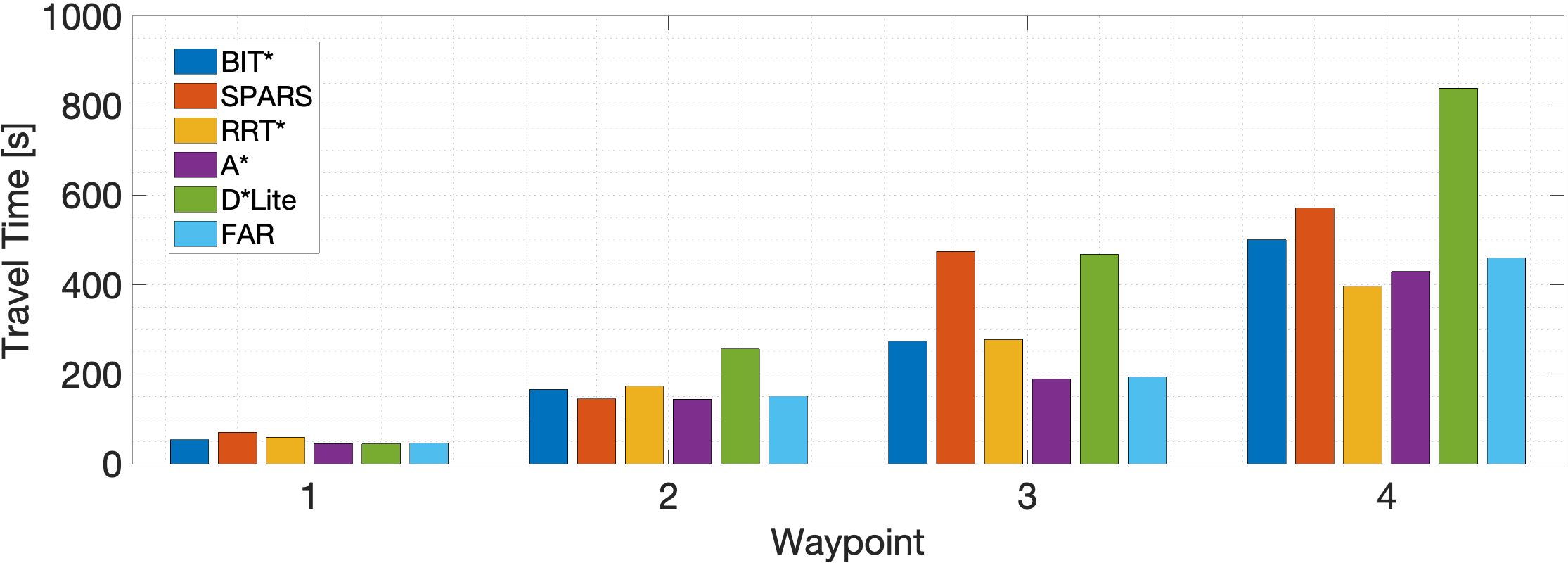}}\\\vspace{-0.1in}
	\subfigure[]{\includegraphics[width=0.79\linewidth]{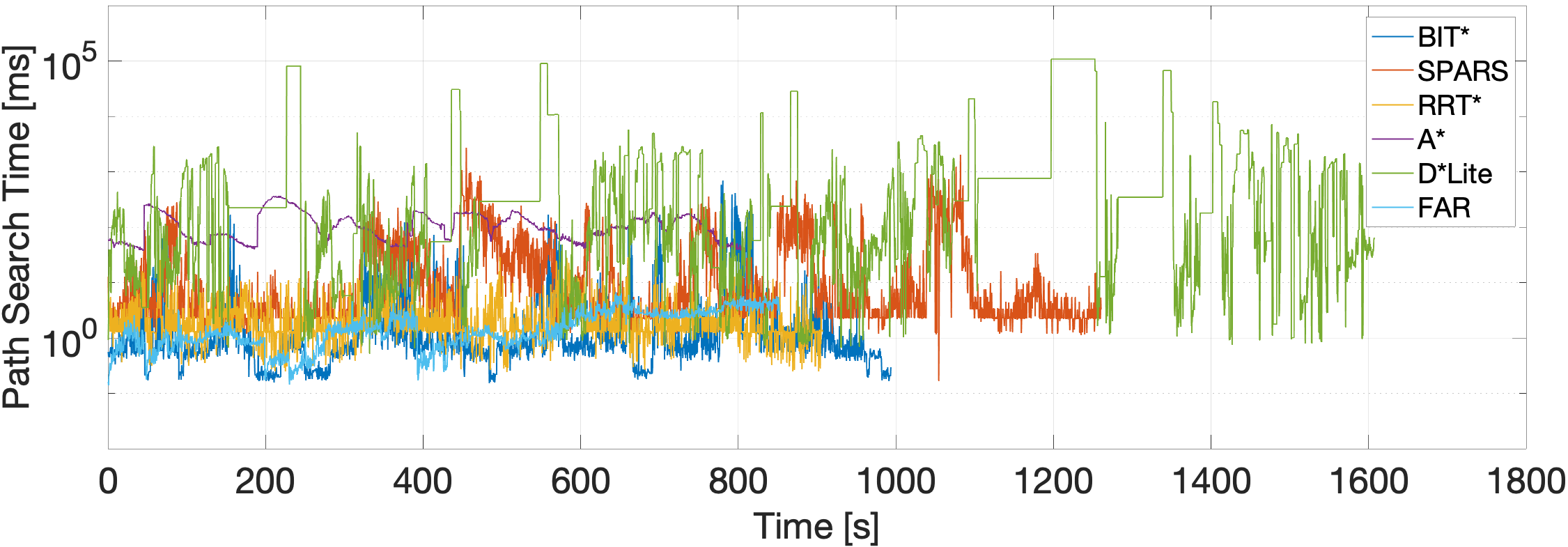}}\\\vspace{-0.1in}
	\subfigure[]{\includegraphics[width=0.79\linewidth]{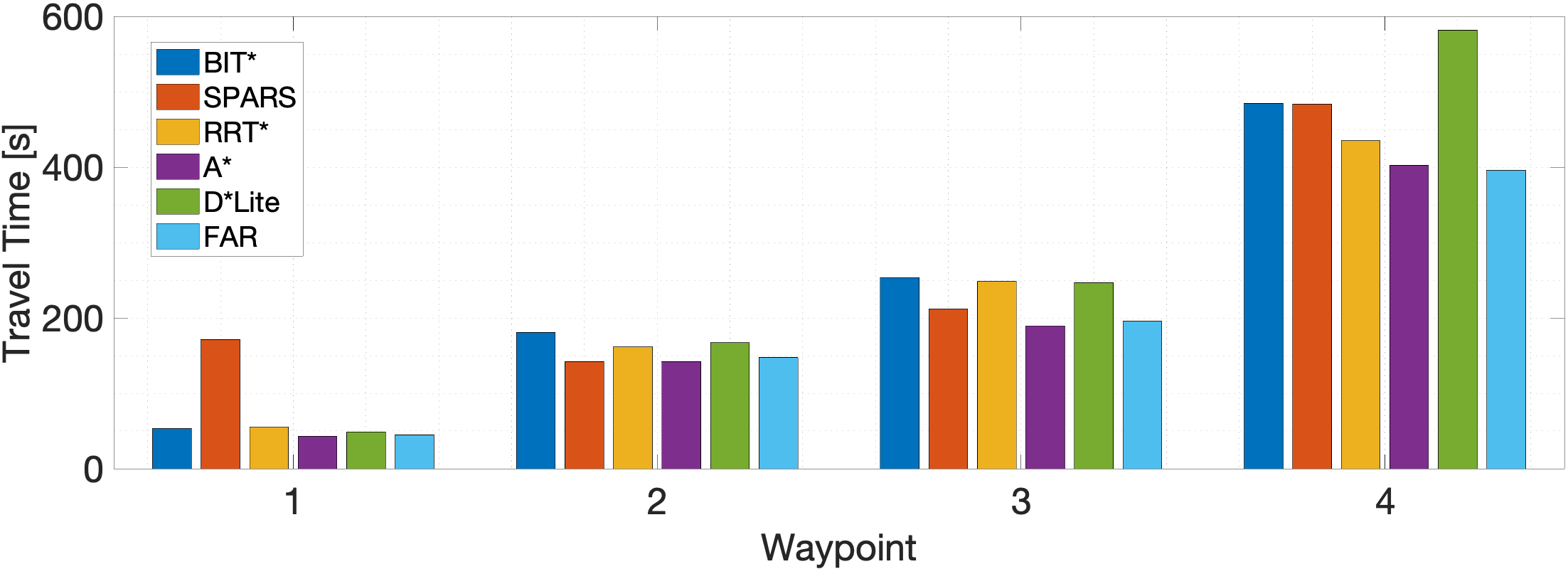}}\\\vspace{-0.1in}
	\subfigure[]{\includegraphics[width=0.79\linewidth]{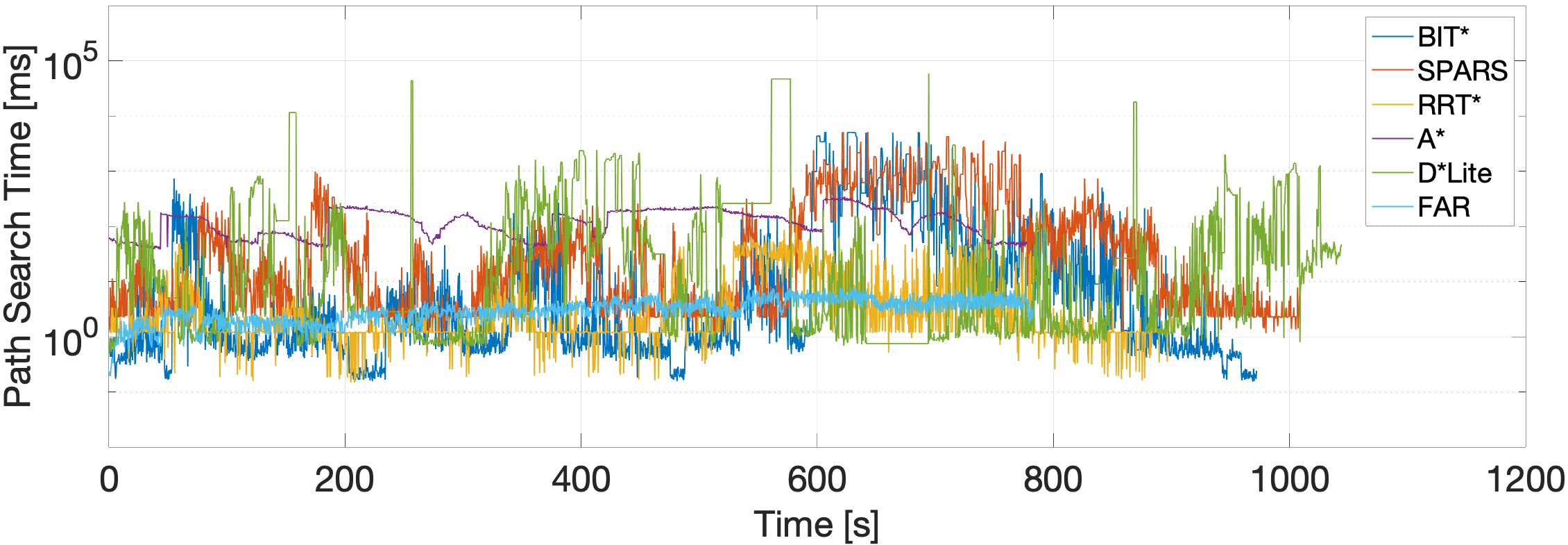}}\\\vspace{-0.1in}
	\caption{Ground vehicle simulation in outdoor campus environment. The vehicle is set to navigate in increasing order from the start (point 0) to point 4 (see labels in (a)). The figure shares the same layout with Fig.~\ref{fig:indoor}.}
	\label{fig:campus}
    \vspace{-0.2in}
\end{figure}

\begin{figure}[t!]
	\centering
	\subfigure[]{\includegraphics[width=0.88\linewidth]{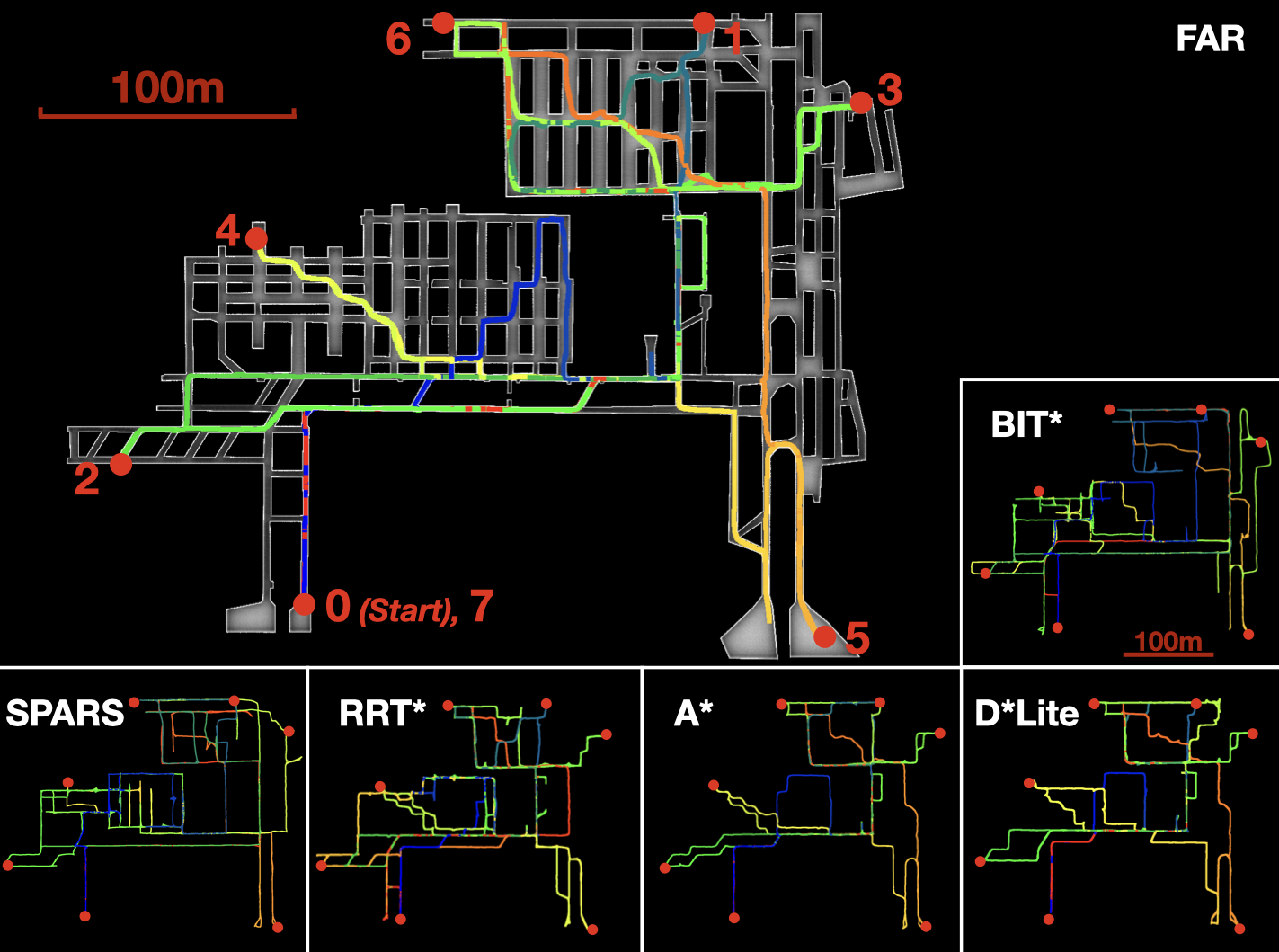}}\\\vspace{-0.09in}
	\subfigure[]{\includegraphics[width=0.79\linewidth]{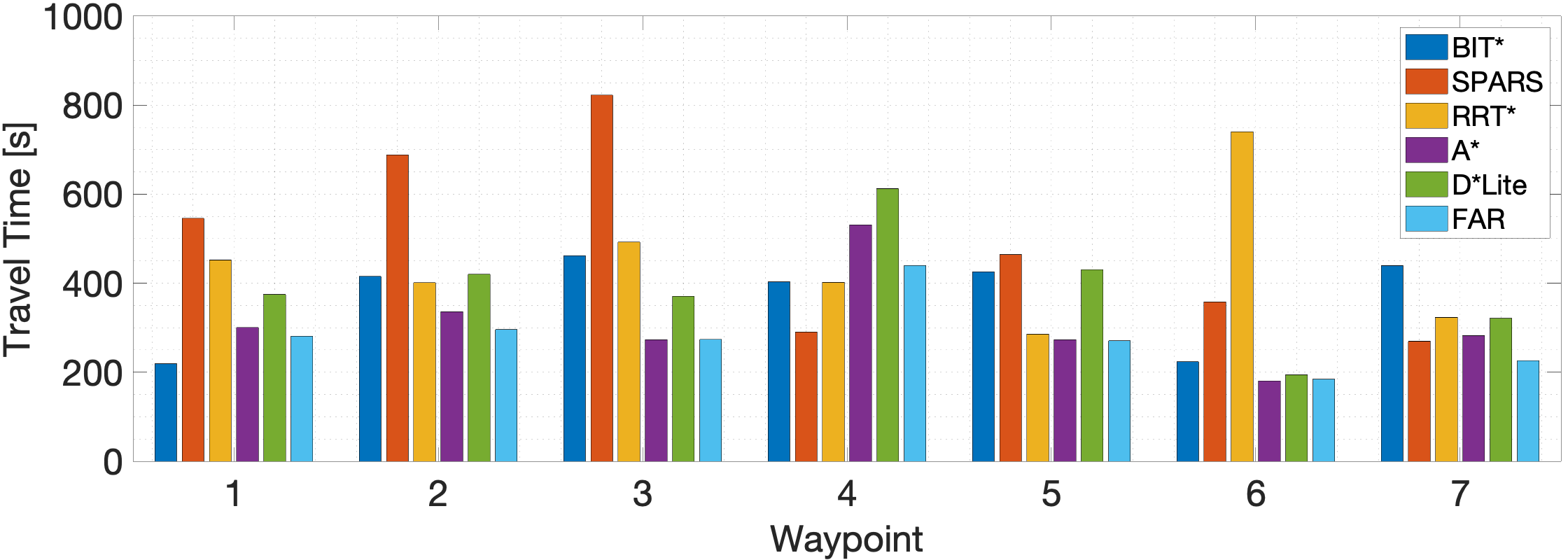}}\\\vspace{-0.1in}
	\subfigure[]{\includegraphics[width=0.79\linewidth]{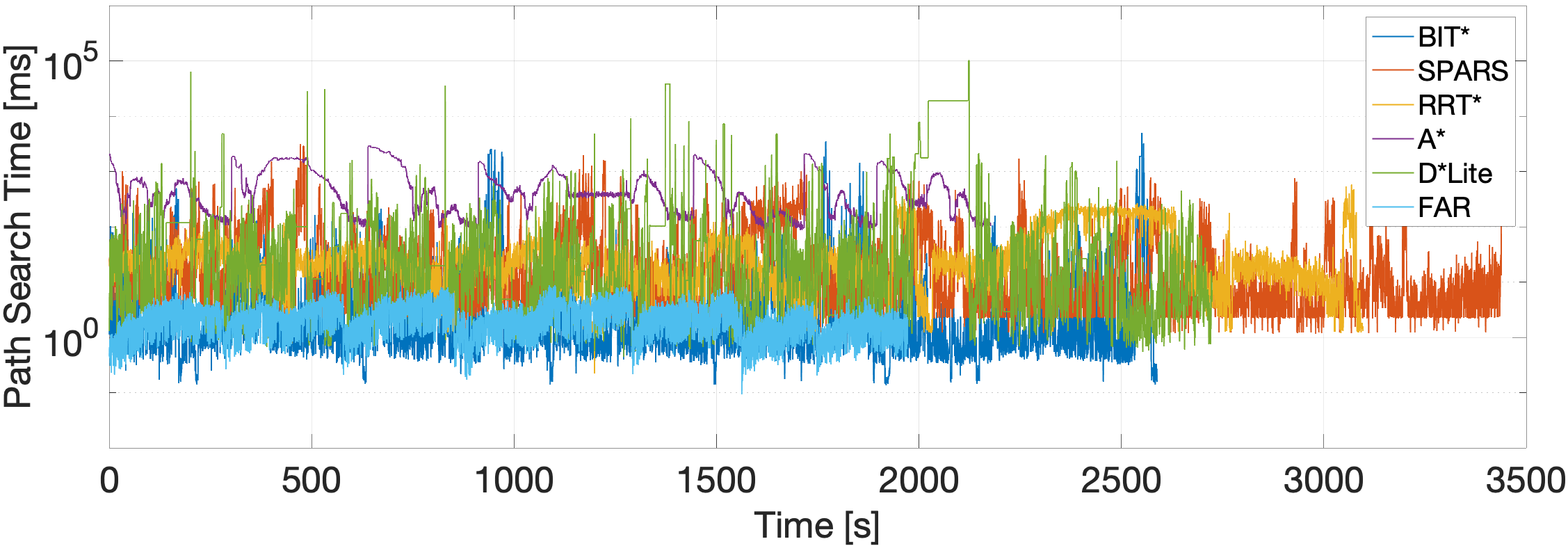}}\\\vspace{-0.1in}
	\subfigure[]{\includegraphics[width=0.79\linewidth]{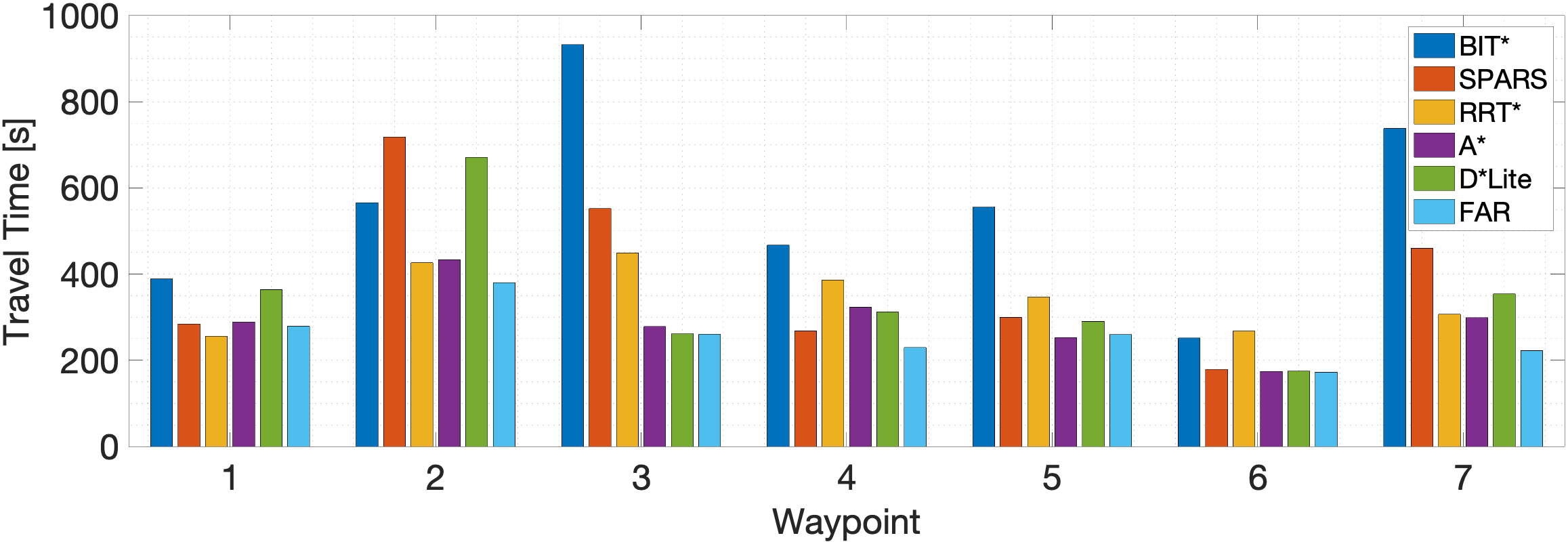}}\\\vspace{-0.1in}
	\subfigure[]{\includegraphics[width=0.79\linewidth]{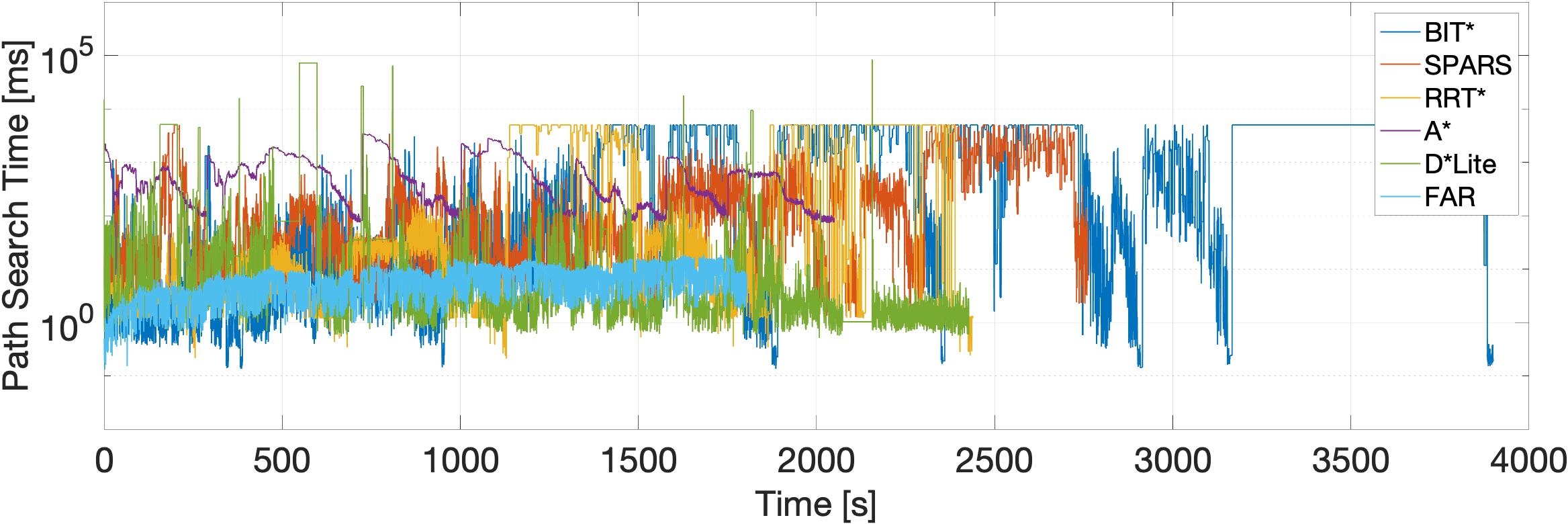}}\\\vspace{-0.1in}
	\caption{Ground vehicle simulation in large-scale, highly convoluted tunnel-network environment. The figure shares the same layout with Fig.~\ref{fig:indoor}.}
	\label{fig:tunnel}
    \vspace{-0.2in}
\end{figure}

\subsection{Ground Vehicle Simulation}

\begin{table}[t]
\renewcommand{\arraystretch}{1.15} \centering{\scriptsize
\caption{Overall travel time in [s] for ground vehicle simulation} 
\label{tab:toa}
\vspace{-0.05in}
\begin{tabular}{||c||c|c|c|c|c|c|c|c|c||}
\hline Test & BIT* & SPARS & RRT* & A* & D* Lite & FAR \\
\hline\hline Indoor (reset) & 1529 & 1506 & 1700 & 1089 & 1080 & \textbf{1032} \\ 
\hline Indoor (accum.) & 1057 & 990 & 1593 & 883 & 921 & \textbf{852} \\ 
\hline Campus (reset) & 994 & 1261 & 907 & \textbf{809} & 1607 & 852 \\ 
\hline Campus (accum.) & 972 & 1009 & 901 & \textbf{777} & 1044 & 784 \\ 
\hline Tunnel (reset) & 2588 & 3439 & 3097 & 2177 & 2724 & \textbf{1972} \\ 
\hline Tunnel (accum.) & 3899 & 2762 & 2439 & 2050 & 2427 & \textbf{1804} \\ 
\hline
\end{tabular}}
\vspace{-0.05in}
\end{table}

\begin{table}[t]
\renewcommand{\arraystretch}{1.15} \centering{\scriptsize
\caption{Average search time in [ms] for ground vehicle simulation} 
\label{tab:searchtime}
\vspace{-0.05in}
\begin{tabular}{||c||c|c|c|c|c|c|c|c|c||}
\hline Test & BIT* & SPARS & RRT* & A* & D* Lite & FAR \\
\hline\hline Indoor (reset) & 20.4 & 27.3 & 39.2 & 59.0 & 28.6 & \textbf{1.59} \\ 
\hline Indoor (accum.) & 203.5 & 132.6 & 294.6 & 58.8 & 27.1 & \textbf{4.11} \\
\hline Campus (reset) & 5.9 & 36.1 & 2.7 & 115.3 & 462.7 & \textbf{1.74} \\ 
\hline Campus (accum.) & 58.3 & 100.0 & 6.5 & 132.2 & 174.2 & \textbf{3.24} \\
\hline Tunnel (reset) & 16.8 & 42.8 & 41.7 & 379.2 & 126.3 & \textbf{2.53} \\ 
\hline Tunnel (accum.) & 392.3 & 179.5 & 169.7 & 394.9 & 94.2 & \textbf{7.37} \\ 
\hline
\end{tabular}}
\vspace{-0.05in}
\end{table}

\begin{table}[!]
\renewcommand{\arraystretch}{1.15} \centering{\scriptsize
\caption{Average processing load [0-100\%] for ground vehicle simulation based on a single CPU thread}
\label{tab:processload}
\vspace{-0.05in}
\begin{tabular}{||c||c|c|c|c|c|c|c|c|c||}
\hline Test & BIT* & SPARS & RRT* & A* & D* Lite & FAR \\
\hline\hline Indoor (reset) & 6.6 & 8.4 & 16.6 & 17.7 & 15.7 & \textbf{6.2} \\ 
\hline Indoor (accum.) & 76.9 & 47.6 & 90.4 & 16.5 & 14.1 & \textbf{11.5} \\ 
\hline Campus (reset) & 3.9 & 12.3 & \textbf{2.9} & 31.3 & 97.2 & 10.2 \\ 
\hline Campus (accum.) & 36.1 & 43.6 & \textbf{3.4} & 35.1 & 77.8 & 18.1 \\
\hline Tunnel (reset) & 9.6 & 15.8 & 15.4 & 89.6 & 78.0 & \textbf{4.3} \\ 
\hline Tunnel (accum.) & 92.1 & 68.9 & 82.6 & 89.9 & 83.9 & \textbf{7.3} \\ 
\hline
\end{tabular}}
\vspace{-0.15in}
\end{table}

The simulated experiments use the same vehicle and sensor configurations as our real ground vehicle platform in Fig.~\ref{fig:vehicles}(a). The speed is set to 2m/s. The experiments are performed in a moderately convoluted indoor environment, a mid-scale outdoor campus environment, and a large-scale, highly convoluted tunnel-network environment. Fig.~\ref{fig:indoor}-\ref{fig:tunnel} show the results from the three environments, respectively. In all three environments, the vehicle is set to navigate through a series of points. Each experiment includes two settings: \textit{unknown} and \textit{partially known}. In the \textit{unknown} environment setting, we reset the planner after the vehicle arrives at each point, such that the environment becomes unknown as the vehicle navigates toward the next point in the series. In the \textit{partially known} environment setting, we do not reset the planner but let it accumulate the environment observations through the entire run, i.e., as the vehicle navigates through the environment, the environment gradually becomes a partially known environment. 

Fig.~\ref{fig:indoor}-\ref{fig:tunnel}(a) show the trajectories from ours and other methods with the \textit{unknown} environment setting. FAR planner is able to produce efficient trajectories similar to A* and D* Lite, while BIT*, SPARS, and RRT* are prone to randomness and often generate back-and-forth patterns along the trajectories. Fig.~\ref{fig:indoor}-\ref{fig:tunnel}(b)-(c) present the travel time to each point and the search time with planner reset. Fig.~\ref{fig:indoor}-\ref{fig:tunnel}(d)-(e) show the same metrics with the \textit{partially known} environment setting, where observations are accumulated along the navigation. Tables~\ref{tab:toa}-\ref{tab:processload} give the overall travel time, the average path search time (planning time), and processing load percentage in terms of the occupancy on a single CPU thread through the entire runs. For FAR planner, the processing load includes the computation for all tasks of polygon extraction, v-graph updates, and path searching.

For search-based methods, A* and D* Lite are known for their resolution completeness in finding the optimal path. However, those methods are difficult to scale as the computational cost increases significantly when environments are large and complex. The long planning time can result in slow responses and a long travel time. Here, the map resolution for both A* and D* is set to 0.2m. Tables~\ref{tab:searchtime} and \ref{tab:processload} show that the search time of A* almost doubles from the indoor to the campus environment, and increases around 3 times from the campus to the tunnel environment, and the processing load increases from 16.5\% in the mid-scale indoor environment to 89.9\% in the large-scale, highly convoluted tunnel environment. D* Lite, on the other hand, reduces the planning time by reusing the state values from the last planning circle. However, when the vehicle reaches a dead-end, many state values become inconsistent and the re-planning takes a large number of iterations for the state values to converge. The searching time of D* Lite increases from 28.6ms in the indoor environment to 462.7ms in the campus environment with the processing load increasing from 15.7\% to 97.2\%.

For sampling-based methods, BIT*, SPARS, and RRT*, their search times are highly inconsistent and 
prone to randomness and often generate back-and-forth patterns along the trajectories which increases the overall travel time. In order to find paths through unknown environments, samples need to be drawn from not only free space but also unknown space. When the environment becomes partially observed, the search time can increase as the connections from unknown space to free space are often blocked by observed obstacles, and thus a feasible path to the goal in unknown space is hard to find. 
As shown in Tables~\ref{tab:searchtime} and \ref{tab:processload}, the search time and processing load for all three sampling-based methods: BIT*, SPARS, and RRT* increase from the \textit{unknown} to \textit{partially known} environment settings. Particularly, in the tunnel environment, the search time of BIT* increases from 16.8ms to 392.3ms while the processing load increases from 9.6\% to 92.1\%, resulting in the overall travel time in the \textit{partially known} environment setting to be 50.7\% longer than the \textit{unknown} environment setting.

As shown in Fig.~\ref{fig:indoor}-\ref{fig:tunnel}(a), FAR planner is able to search for optimal paths on the graph and generate efficient trajectories through unknown environments to the goals. Table~\ref{tab:toa} shows that in the indoor and tunnel environments, FAR planner reduces the travel time up to 12.0\% from A* and up to 47.0\% from D* Lite. In the outdoor campus environment, A* marginally surpasses FAR planner in terms of the travel time. 
However, FAR planner consumes only around half of the processing load and is nearly two orders of magnitude faster than A* in re-planning. 

Compared to the sampling-based planners, FAR planner surpasses BIT*, SPARS, and RRT* in terms of travel time as well as consistency of planning time and processing load in different environment settings. In the tunnel environment, FAR planner finishes the run more than 23.8\% faster than BIT*, more than 34.7\% faster than SPARS, and more than 26.0\% faster than RRT*. Further, with the two-layered v-graph framework, FAR planner maintains a low processing load in both the \textit{unknown} and \textit{partially known} environment settings with an average processing load less than 20\% of a single CPU thread, shown in Table~\ref{tab:processload}. In addition, FAR planner demonstrates fast re-planning with average search time less than 10ms in all experiments, shown in Table~\ref{tab:searchtime}.

\begin{figure}[t!]
	\centering
	\subfigure[]{\includegraphics[width=0.95\linewidth]{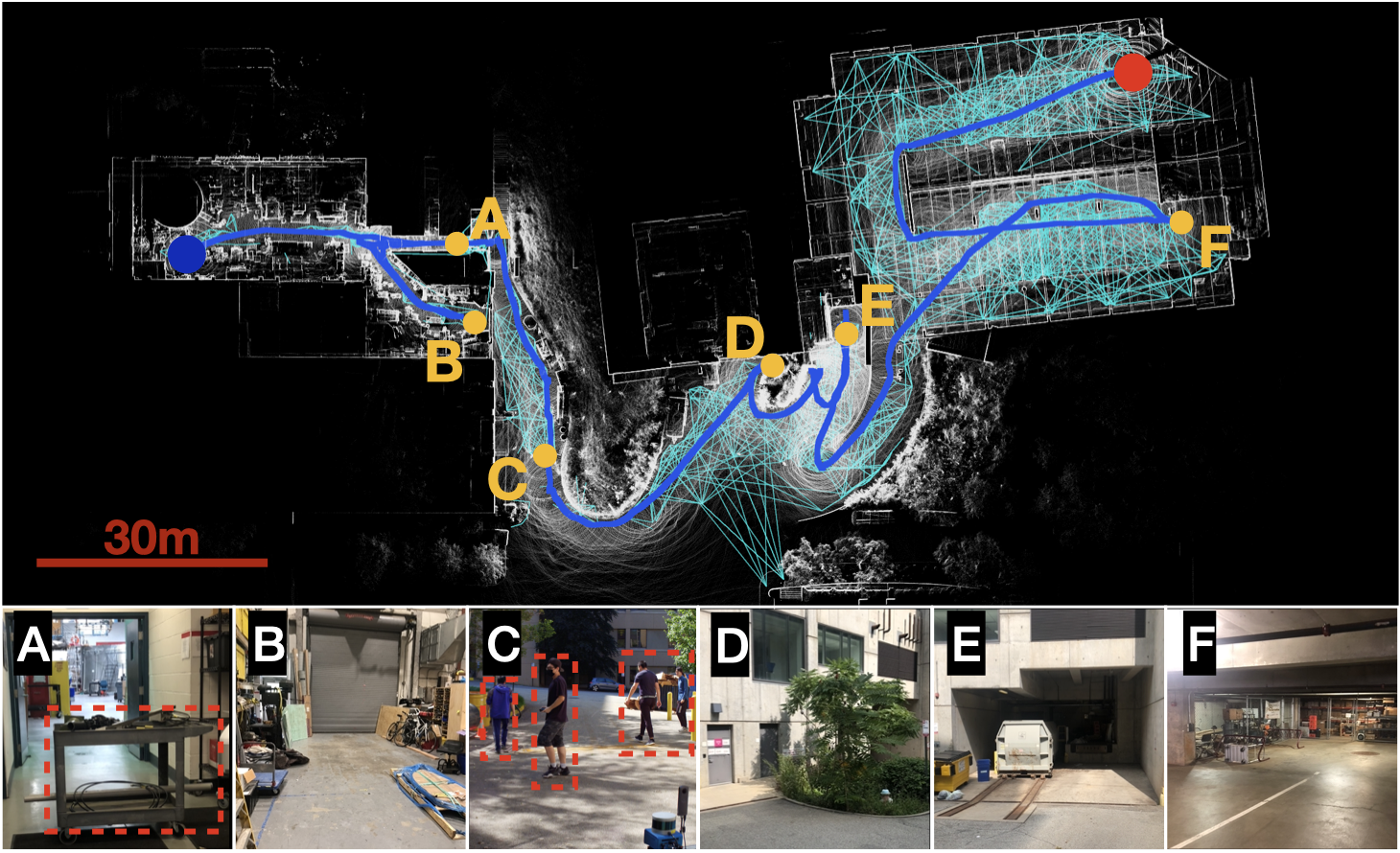}}\\\vspace{-0.1in}
	\subfigure[]{\includegraphics[width=0.95\linewidth]{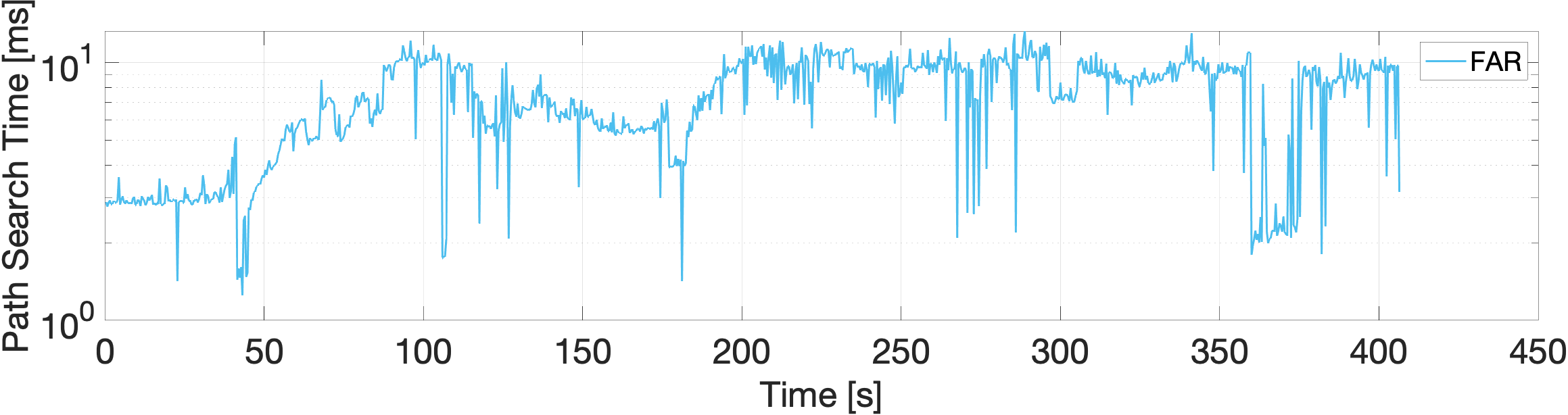}}\\\vspace{-0.05in}
	\caption{Ground vehicle physical experiment. In (a), the blue curve is the vehicle trajectory staring at the blue dot and ending at the red dot. The vehicle starts from the inside of a building, navigates to the outside, and reaches goal in a garage building. The environment is unknown. The visibility edges (cyan) are developed during the navigation. The small images on the bottom are taken from A to F as labeled on the trajectory. B, D, E, and F are four dead-ends where the vehicle re-routes. At A, a cart in a corridor blocks the path forcing the vehicle to choose another way and get into a dead-end at B. On the way back, the cart is removed and the vehicle navigates through the corridor to the outside. The area around C involves dynamic obstacles. (b) shows the search time through the run.}
	\label{fig:wheelchair}
	\vspace{-0.15in}
\end{figure}

\begin{figure}[b!]
	\vspace{-0.1in}
    \centering
    \includegraphics[width=0.97\linewidth]{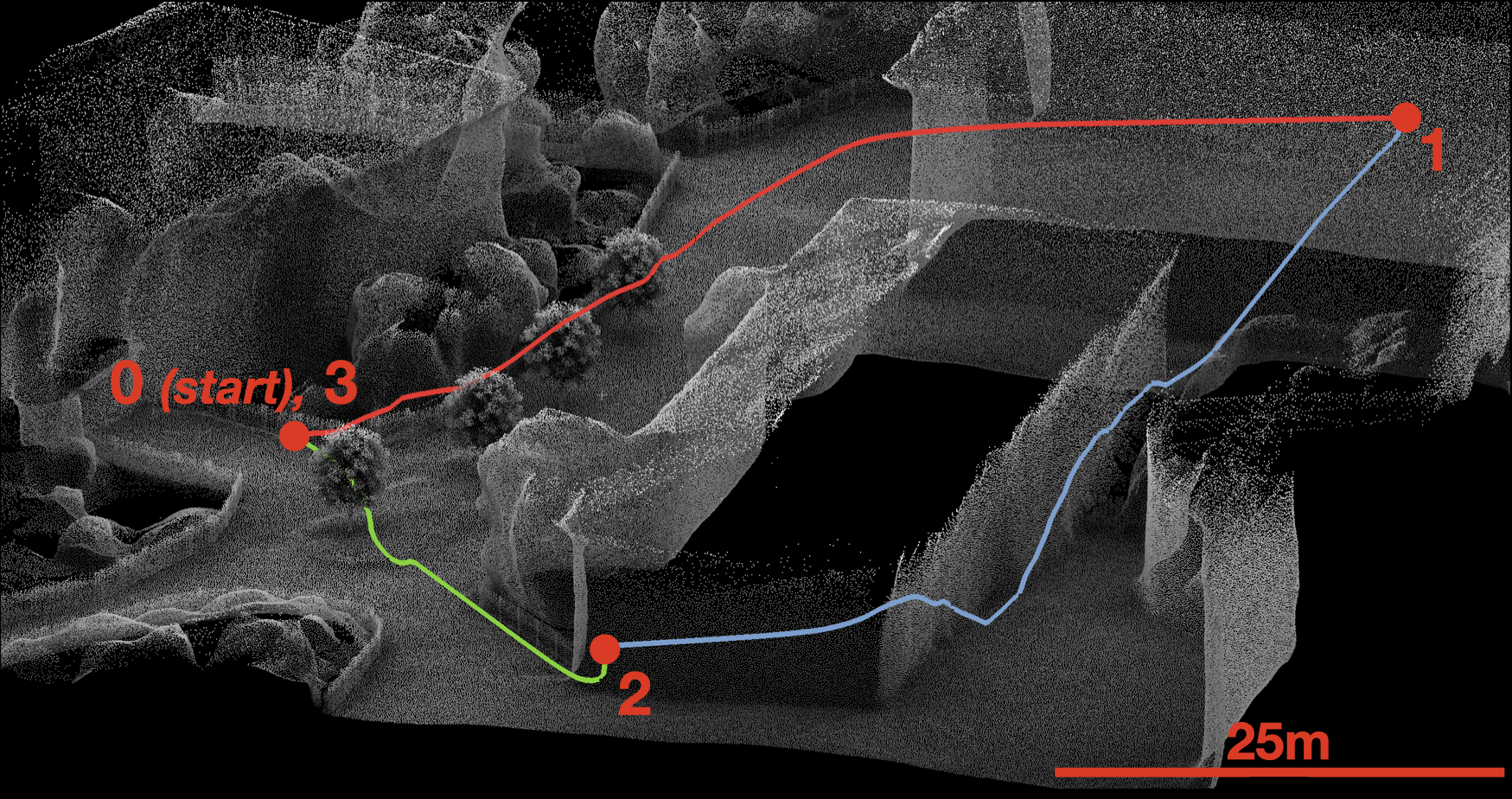}
	\vspace{-0.05in}
    \caption{A preliminary result of aerial vehicle simulation in campus environment. Similar to the ground vehicle simulation experiments, the vehicle is set to navigate in increasing order from the start to point 3 as point able and the trajectories shown in the figure. The planner is reset after the vehicle arrives at each point.}
	\label{fig:aerial}
\end{figure}

\subsection{Ground Vehicle Physical Experiment}

The physical experiment uses the ground vehicle platform in Fig.~\ref{fig:vehicles}(a) with the speed set to 1.5m/s. As shown in Fig.~\ref{fig:wheelchair}, the vehicle starts from the inside of a building, navigates to the outside, and reaches the goal in a garage building. The environment is unknown beforehand. The vehicle gets into four dead-ends and then re-routes. A cart initially blocks the vehicle path forcing it to choose another way that leads to a dead-end. While the vehicle navigates out of the dead-end, the cart is removed allowing the vehicle to pass through. In the outdoor area, pedestrians are present as dynamic obstacles. FAR planner disconnects the visibility edges blocked by the pedestrians and later on reconnects them after the pedestrians move away. Fig.~\ref{fig:aerial}(b) presents the search time through the run (average 7.32ms). The vehicle drivers 388m in 406s.

FAR planner is used by the CMU-OSU team as the main route planner in attending DARPA Subterranean Challenge. In the final competition, the team received a ``Most Sectors Explored Award" by conducting the most complete traversing and mapping across the site (26 out of 28 sectors).

\subsection{Aerial Vehicle Simulation}

We use the simulated aerial platform in Fig.~\ref{fig:vehicles}(b) with the speed set to 4m/s. The 3D version of FAR planner uses multi-layer polygons. The vertical resolution between the polygon layers is 1m. Fig.~\ref{fig:aerial} shows a preliminary result in simulation using an environment based on the university campus. The vehicle follows a series of points and navigates as in an unknown environment. Fig.~\ref{fig:aerial} presents the goal points and fly trajectory. Overall, the vehicle flies through a 210m trajectory in 57s.


\section{Conclusion}

We present a planning framework based on visibility graph. The method is capable of efficiently handling unknown and partially known environments. The method involves registering obstacle data into polygons and using a two-layer data structure in dynamically updating the visibility graph for a low computational cost. The path search is conducted by propagating through the visibility graph, finding paths at low latency. We benchmark the performance of the state-of-the-art sampling-based methods: BIT*, SPARS, and the classic RRT*, as well as search-based methods: A* and D* Lite, in unknown and partially known environments, Further, we evaluate our method in both simulated and real-world environments with ground vehicles and produce preliminary results using a multi-layer 3D v-graph for aerial vehicles. The comparison results indicate that our method is efficient in both unknown and partially known environments with a constant low processing load and uses significantly less planning time compared to the state-of-the-art.

\bibliographystyle{IEEEtran}
\bibliography{references}

\begin{thebibliography}{10}
\providecommand{\url}[1]{#1}
\csname url@samestyle\endcsname
\providecommand{\newblock}{\relax}
\providecommand{\bibinfo}[2]{#2}
\providecommand{\BIBentrySTDinterwordspacing}{\spaceskip=0pt\relax}
\providecommand{\BIBentryALTinterwordstretchfactor}{4}
\providecommand{\BIBentryALTinterwordspacing}{\spaceskip=\fontdimen2\font plus
\BIBentryALTinterwordstretchfactor\fontdimen3\font minus
  \fontdimen4\font\relax}
\providecommand{\BIBforeignlanguage}[2]{{%
\expandafter\ifx\csname l@#1\endcsname\relax
\typeout{** WARNING: IEEEtran.bst: No hyphenation pattern has been}%
\typeout{** loaded for the language `#1'. Using the pattern for}%
\typeout{** the default language instead.}%
\else
\language=\csname l@#1\endcsname
\fi
#2}}
\providecommand{\BIBdecl}{\relax}
\BIBdecl

\bibitem{10.1145/359156.359164}
T.~Lozano-P\'{e}rez and M.~A. Wesley, ``An algorithm for planning
  collision-free paths among polyhedral obstacles,'' \emph{Communications of
  the ACM}, vol.~22, no.~10, p. 560–570, 1979.

\bibitem{Kitzinger2003TheVG}
J.~Kitzinger, ``The visibility graph among polygonal obstacles: a comparison of
  algorithms,'' 2003.

\bibitem{Chao2022DevEnv}
C.~Cao, H.~Zhu, F.~Yang, Y.~Xia, H.~Choset, J.~Oh, and J.~Zhang, ``Autonomous
  exploration development environment and the planning algorithms,'' in
  \emph{IEEE International Conference on Robotics and Automation},
  Philadelphia, PA, May 2022.

\bibitem{lavalle2001rapidly}
S.~M. LaValle, J.~J. Kuffner, B.~Donald \emph{et~al.}, ``Rapidly-exploring
  random trees: Progress and prospects,'' \emph{Algorithmic and computational
  robotics: new directions}, vol.~5, pp. 293--308, 2001.

\bibitem{karaman2011sampling}
S.~Karaman and E.~Frazzoli, ``Sampling-based algorithms for optimal motion
  planning,'' \emph{The International Journal of Robotics Research}, vol.~30,
  no.~7, pp. 846--894, 2011.

\bibitem{844730}
J.~Kuffner and S.~LaValle, ``{RRT}-connect: An efficient approach to
  single-query path planning,'' in \emph{IEEE International Conference on
  Robotics and Automation}, 2000, pp. 995--1001.

\bibitem{6942976}
J.~D. Gammell, S.~S. Srinivasa, and T.~D. Barfoot, ``{Informed RRT*}: Optimal
  sampling-based path planning focused via direct sampling of an admissible
  ellipsoidal heuristic,'' in \emph{IEEE/RSJ International Conference on
  Intelligent Robots and Systems}, 2014, pp. 2997--3004.

\bibitem{bitrrt2038457}
J.~Gammell, S.~Srinivasa, and T.~Barfoot, ``{Batch Informed Trees (BIT*)}:
  Sampling-based optimal planning via the heuristically guided search of
  implicit random geometric graphs,'' in \emph{IEEE International Conference on
  Robotics and Automation}, 06 2015, pp. 3067--3074.

\bibitem{508439}
L.~Kavraki, P.~Svestka, J.-C. Latombe, and M.~Overmars, ``Probabilistic
  roadmaps for path planning in high-dimensional configuration spaces,''
  \emph{IEEE Transactions on Robotics and Automation}, vol.~12, no.~4, pp.
  566--580, 1996.

\bibitem{844107}
R.~Bohlin and L.~Kavraki, ``Path planning using lazy {PRM},'' in \emph{IEEE
  International Conference on Robotics and Automation}, 2000, pp. 521--528.

\bibitem{6631156}
A.~Dobson and K.~E. Bekris, ``Improving sparse roadmap spanners,'' in
  \emph{2013 IEEE International Conference on Robotics and Automation}, 2013,
  pp. 4106--4111.

\bibitem{4269896}
Y.~Tian, L.~Yan, G.-Y. Park, S.-H. Yang, Y.-S. Kim, S.-R. Lee, and C.-Y. Lee,
  ``Application of {RRT}-based local path planning algorithm in unknown
  environment,'' in \emph{International Symposium on Computational Intelligence
  in Robotics and Automation}, 2007, pp. 456--460.

\bibitem{1242258}
C.~Lanzoni, A.~Sanchez, and R.~Zapata, ``Sensor-based motion planning for
  car-like mobile robots in unknown environments,'' in \emph{IEEE International
  Conference on Robotics and Automation}, 2003, pp. 4258--4263.

\bibitem{1641879}
D.~Ferguson, N.~Kalra, and A.~Stentz, ``Replanning with {RRT}s,'' in \emph{IEEE
  International Conference on Robotics and Automation}, 2006, pp. 1243--1248.

\bibitem{dijkstra1959note}
E.~W. Dijkstra, ``A note on two problems in connexion with graphs,''
  \emph{Numerische mathematik}, vol.~1, no.~1, pp. 269--271, 1959.

\bibitem{Hart1968}
P.~Hart, N.~Nilsson, and B.~Raphael, ``A formal basis for the heuristic
  determination of minimum cost paths,'' \emph{{IEEE} Transactions on Systems
  Science and Cybernetics}, vol.~4, no.~2, pp. 100--107, 1968.

\bibitem{stentz1997optimal}
A.~Stentz, ``Optimal and efficient path planning for partially known
  environments,'' in \emph{Intelligent Unmanned Ground Vehicles}, 1997, pp.
  203--220.

\bibitem{koenig2005fast}
S.~Koenig and M.~Likhachev, ``Fast replanning for navigation in unknown
  terrain,'' \emph{IEEE Transactions on Robotics}, vol.~21, no.~3, pp.
  354--363, 2005.

\bibitem{7926533}
A.~T. Le, M.~Q. Bui, T.~D. Le, and N.~Peter, ``D* {L}ite with reset: Improved
  version of {D}* {L}ite for complex environment,'' in \emph{IEEE International
  Conference on Robotic Computing}, 2017, pp. 160--163.

\bibitem{5679403}
S.~C. Yun, V.~Ganapathy, and T.~W. Chien, ``Enhanced {D}* {L}ite algorithm for
  mobile robot navigation,'' in \emph{IEEE Symposium on Industrial Electronics
  and Applications}, 2010, pp. 545--550.

\bibitem{Bayesian43242}
C.~Richter and N.~Roy, \emph{Bayesian Learning for Safe High-Speed Navigation
  in Unknown Environments}, 01 2018, pp. 325--341.

\bibitem{learnNav24i3204}
J.~Zeng, R.~Ju, L.~Qin, Y.~Hu, Q.~Yin, and C.~Hu, ``Navigation in unknown
  dynamic environments based on deep reinforcement learning,'' \emph{Sensors},
  vol.~19, p. 3837, 09 2019.

\bibitem{9066637}
X.~Guo and Y.~Fang, ``Learning to navigate in unknown environments based on
  {GMRP-N},'' in \emph{IEEE Annual International Conference on CYBER Technology
  in Automation, Control, and Intelligent Systems}, 2019, pp. 1453--1458.

\bibitem{1087133}
B.~Oommen, S.~Iyengar, N.~Rao, and R.~Kashyap, ``Robot navigation in unknown
  terrains using learned visibility graphs. part i: The disjoint convex
  obstacle case,'' \emph{IEEE Journal on Robotics and Automation}, vol.~3,
  no.~6, pp. 672--681, 1987.

\bibitem{384257}
N.~Rao, ``Robot navigation in unknown generalized polygonal terrains using
  vision sensors,'' \emph{IEEE Transactions on Systems, Man, and Cybernetics},
  vol.~25, no.~6, pp. 947--962, 1995.

\bibitem{1642054}
D.~Wooden and M.~Egerstedt, ``Oriented visibility graphs: low-complexity
  planning in real-time environments,'' in \emph{Proceedings 2006 IEEE
  International Conference on Robotics and Automation, 2006. ICRA 2006.}, 2006,
  pp. 2354--2359.

\bibitem{5967147}
H.~Kaluder, M.~Brezak, and I.~Petrovic, ``A visibility graph based method for
  path planning in dynamic environments,'' in \emph{The 34th International
  Convention MIPRO}, 2011, pp. 717--721.

\bibitem{Visibility118.123}
M.~El~Khaili, ``Visibility graph for path planning in the presence of moving
  obstacles,'' \emph{Engineering Science and Technology an International
  Journal}, vol.~4, pp. 118--123, 09 2014.

\bibitem{Suzuki1985TopologicalSA}
S.~Suzuki and K.~Abe, ``Topological structural analysis of digitized binary
  images by border following,'' \emph{Computer Vision, Graphics, and Image
  Processing}, vol.~30, pp. 32--46, 1985.

\bibitem{Douglas1973ALGORITHMSFT}
D.~H. Douglas and T.~K. Peucker, ``Algorithms for the reduction of the number
  of points required to represent a digitized line or its caricature,''
  \emph{Cartographica: The International Journal for Geographic Information and
  Geovisualization}, vol.~10, pp. 112--122, 1973.

\bibitem{Lee78}
D.-T. Lee, ``Proximity and reachability in the plane.'' Ph.D. dissertation,
  USA, 1978.

\bibitem{andersen2008modern}
R.~Andersen and S.~Publications, \emph{Modern Methods for Robust
  Regression}.\hskip 1em plus 0.5em minus 0.4em\relax SAGE Publications, 2008.

\bibitem{zhang2018slam}
J.~Zhang and S.~Singh, ``Laser-visual-inertial odometry and mapping with high
  robustness and low drift,'' \emph{Journal of Field Robotics}, vol.~35, no.~8,
  pp. 1242--1264, 2018.

\bibitem{zhang2020avoidance}
J.~Zhang, C.~Hu, R.~G. Chadha, and S.~Singh, ``Falco: Fast likelihood-based
  collision avoidance with extension to human-guided navigation,''
  \emph{Journal of Field Robotics}, vol.~37, no.~8, pp. 1300--1313, 2020.

\end{thebibliography}

\end{document}